\pdfoutput=1
\documentclass{article}

     \PassOptionsToPackage{numbers, compress}{natbib}

\usepackage[preprint]{neurips_2022}

\usepackage[utf8]{inputenc} %
\usepackage[T1]{fontenc}    %
\usepackage{hyperref}       %
\usepackage{url}            %
\usepackage{booktabs}       %
\usepackage{amsfonts}       %
\usepackage{nicefrac}       %
\usepackage{microtype}      %
\usepackage{xcolor}         %
\usepackage[ruled]{algorithm2e}
\usepackage[pdftex]{graphicx}
\usepackage{multicol}
\usepackage{subcaption}

\usepackage[shortcuts]{extdash}  %

\newcommand \dsame {DSA\=/ME}

\newcommand \cmame {CMA\=/ME}
\newcommand \mapelites {MAP\=/Elites}
\newcommand \dsage {DSAGE}
\newcommand \dsageoa {DSAGE\=/Only~Anc}
\newcommand \dsageod {DSAGE\=/Only~Down}
\newcommand \dsageb {DSAGE~Basic}

\usepackage{xcolor}

\newcommand{\xxnote}[3]{}
\ifx\hidenotes\undefined
  \renewcommand{\xxnote}[3]{\color{#2}{(#1: #3)}}
\fi

\definecolor{lightgrey}{rgb}{0.9, 0.9, 0.9}

\newcommand{\eref}[1]{Eq.~\ref{#1}}
\newcommand{\sref}[1]{Sec.~\ref{#1}}

\newcommand{\fref}[1]{Fig.~\ref{#1}}
\newcommand{\tref}[1]{Table~\ref{#1}}
\newcommand{\aref}[1]{Algorithm~\ref{#1}}

\usepackage{booktabs}
\usepackage{multirow}
\usepackage{array}
\newcolumntype{L}[1]
  {>{\raggedright\let\newline\\\arraybackslash\hspace{0pt}}m{#1}}
\newcolumntype{C}[1]
  {>{\centering\let\newline\\\arraybackslash\hspace{0pt}}m{#1}}
\newcolumntype{R}[1]
  {>{\raggedleft\let\newline\\\arraybackslash\hspace{0pt}}m{#1}}

\usepackage{amsmath,amsfonts,bm}

\def\eqref#1{equation~\ref{#1}}

\def\1{\bm{1}}

\def\vone{{\bm{1}}}

\def\vtheta{{\bm{\theta}}}

\def\vm{{\bm{m}}}

\DeclareMathAlphabet{\mathsfit}{\encodingdefault}{\sfdefault}{m}{sl}
\SetMathAlphabet{\mathsfit}{bold}{\encodingdefault}{\sfdefault}{bx}{n}

\usepackage{tikz}
\usetikzlibrary{decorations.pathreplacing,calc}
\newcommand{\tikzmark}[1]{\tikz[overlay,remember picture] \node (#1) {};}

\definecolor{algo_imp}{RGB}{221,179,16}
\definecolor{algo_exploit}{RGB}{64,83,211}
\definecolor{algo_sim}{RGB}{181,29,20}

\newcommand*{\AddNote}[5]{%
    \begin{tikzpicture}[overlay, remember picture]
        \draw [decoration={brace,amplitude=0.5em},decorate,ultra thick,{#4}]
            ($(#3)!(#1.north)!($(#3)-(0,1)$)$) --  
            ($(#3)!(#2.south)!($(#3)-(0,1)$)$)
                node [align=left, text width=3cm, xshift=0.2cm, pos=0.5, anchor=west] {#5};
    \end{tikzpicture}
}%

\newcommand*{\AddNoteHacked}[5]{%
    \begin{tikzpicture}[overlay, remember picture]
        \draw [decoration={brace,amplitude=0.5em},decorate,ultra thick,{#4}]
            ($(#3)!(#1.north)!($(#3)-(0,1)$)$) --  
            ($(#3)!($(#2.south) + (0,0.36)$)!($(#3)-(0,1)$)$)
                node [align=left, text width=3cm, xshift=0.2cm, pos=0.5, anchor=west] {#5};
    \end{tikzpicture}
}%

\newcommand*{\AddNoteHackedTwo}[5]{%
    \begin{tikzpicture}[overlay, remember picture]
        \draw [decoration={brace,amplitude=0.5em},decorate,ultra thick,{#4}]
            ($(#3)!(#1.north)!($(#3)-(0,1)$)$) --  
            ($(#3)!($(#2.south) + (0,-0.36)$)!($(#3)-(0,1)$)$)
                node [align=left, text width=3cm, xshift=0.2cm, pos=0.5, anchor=west] {#5};
    \end{tikzpicture}
}%

\title{Deep Surrogate Assisted Generation of Environments}

\author{%
    Varun Bhatt\thanks{Equal contribution} \\
    University of Southern California\\
    Los Angeles, CA\\
    \texttt{vsbhatt@usc.edu} \\
    \And
    Bryon Tjanaka\footnotemark[1] \\
    University of Southern California\\
    Los Angeles, CA\\
    \texttt{tjanaka@usc.edu} \\
    \AND
    Matthew C.~Fontaine\footnotemark[1] \\
    University of Southern California\\
    Los Angeles, CA\\
    \texttt{mfontain@usc.edu} \\
    \And
    Stefanos Nikolaidis \\
    University of Southern California\\
    Los Angeles, CA\\
    \texttt{nikolaid@usc.edu} \\
}

\begin{document}

\maketitle

\begin{abstract}

Recent progress in reinforcement learning (RL) has started producing generally capable agents that can solve a distribution of complex environments. These agents are typically tested on fixed, human-authored environments. On the other hand, quality diversity (QD) optimization has been proven to be an effective component of environment generation algorithms, which can generate collections of high-quality environments that are diverse in the resulting agent behaviors. However, these algorithms require potentially expensive simulations of agents on newly generated environments. We propose Deep Surrogate Assisted Generation of Environments (DSAGE), a sample-efficient QD environment generation algorithm that maintains a deep surrogate model for predicting agent behaviors in new environments. Results in two benchmark domains show that DSAGE significantly outperforms existing QD environment generation algorithms in discovering collections of environments that elicit diverse behaviors of a state-of-the-art RL agent and a planning agent. Our source code and videos are available at \url{https://dsagepaper.github.io/}

\end{abstract}

\section{Introduction}

We present an efficient method of automatically generating a collection of environments that elicit diverse agent behaviors. As a motivating example, consider deploying a robot agent at scale in a variety of home environments. The robot should generalize by performing robustly not only in test homes, but in any end user's home. To validate agent generalization, the test environments should have good coverage for the robot agent.
However, obtaining such coverage may be difficult, as the generated environments would depend on the application domain, e.g. kitchen or living room, and on the specific agent we want to test, since different agents exhibit different behaviors.

To enable generalization of autonomous agents to new environments with differing levels of complexity, previous work on open-ended learning~\citep{wang2019paired,wang2020enhanced} has integrated the environment generation and the agent training processes. The interplay between the two processes acts as a natural curriculum for the agents to learn robust skills that generalize to new, unseen environments~\cite{dennis2020emergent,parker2022evolving,dharna2022transfer}. The performance of these agents has been evaluated either in environments from the training distribution~\citep{wang2019paired,wang2020enhanced,dharna2022transfer} or in suites of manually authored environments~\cite{dennis2020emergent,jiang2021replay,parker2022evolving}.

As a step towards testing generalizable agents, there has been increasing interest in competitions~\citep{perez2016general,hambro2022insights} that require agents to generalize to new game layouts. Despite the recent progress of deep learning agents in fixed game domains, e.g. in Chess~\citep{silver2018general}, Go~\citep{silver2016mastering}, Starcraft~\citep{vinyals2019alphastar}, and Poker~\citep{moravvcik2017deepstack,brown2019superhuman}, it has been rule-based agents that have succeeded in these competitions~\citep{hambro2022insights}.  Such competitions also rely on manually authored game levels as a test set, handcrafted by a human designer. 

While manually authored environments are important for standardized testing, creating these environments can be tedious and time-consuming. Additionally, manually authored test suites are often insufficient for eliciting the diverse range of possible agent behaviors. Instead, we would like an interactive test set that proposes an environment, observes the agent's performance and behavior, and then proposes new environments that diversify the agent behaviors, based on what the system has learned from previous execution traces of the agent.

\begin{figure}
  \centering
  \includegraphics[width=0.985\textwidth]{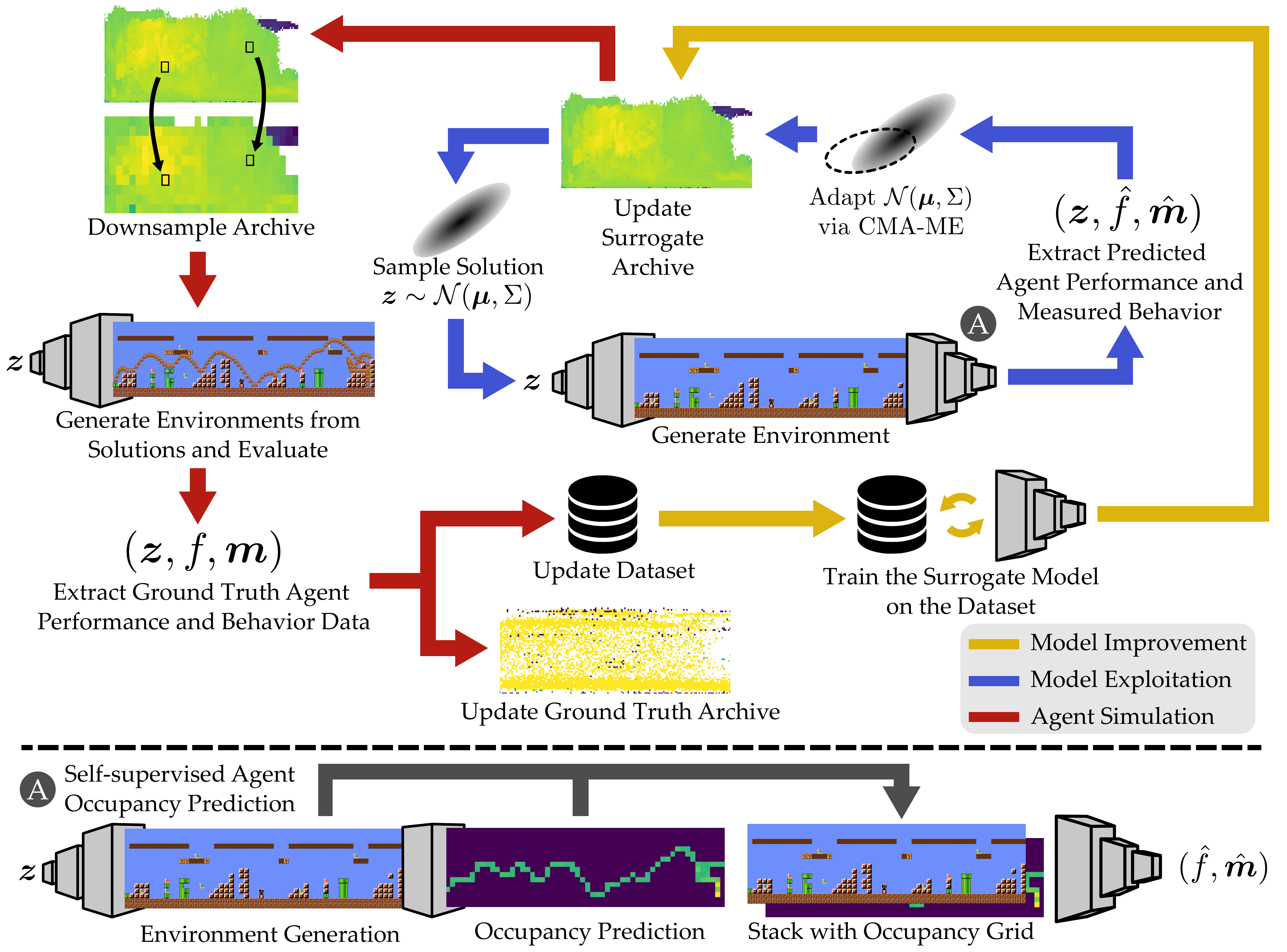}
  \caption{An overview of the Deep Surrogate Assisted Generation of Environments (\dsage{}) algorithm.
  The algorithm begins by generating and evaluating random environments to initialize the dataset and the surrogate model (not shown in the figure).
  An archive of solutions is generated by exploiting a deep surrogate model (\textbf{\color{algo_exploit}blue arrows}) with a QD optimizer, e.g., \cmame{}~\citep{fontaine2020covariance}.
  A subset of solutions from this archive are chosen by downsampling and evaluated by generating the corresponding environment and simulating an agent (\textbf{\color{algo_sim}red arrows}). 
  The surrogate model is then trained on the data from the simulations (\textbf{\color{algo_imp}yellow arrows}).
  While the images show Mario levels, the algorithm structure is similar for mazes.}
  \label{fig:alg}
\end{figure}

To address collecting environments with diverse agent behaviors, prior work frames the problem as a quality diversity (QD) problem~\citep{fontaine2021importance, fontaine2021illuminating, fontaine:sa_rss2021}. A QD problem consists of an objective function, e.g. whether the agent can solve the environment, and measure functions, e.g. how long the agent takes to complete their task. The measure functions quantify the behavior we would like to vary in the agent, allowing practitioners to specify the case coverage they would like to see in the domain they are testing. While QD algorithms can generate diverse collections of environments, they require a large number of environment evaluations to produce the collection, and each of these evaluations requires multiple time-consuming simulated executions of potentially stochastic agent policies.

We study how \textit{deep surrogate models that predict agent performance can accelerate the generation of environments that are diverse in agent behaviors}.  We draw upon insights from model-based quality diversity algorithms that have been previously shown to improve sample efficiency in design optimization~\citep{gaier2018dataefficient} and Hearthstone deckbuilding~\citep{zhang2021deep}. Environments present a much more complex prediction task because the evaluation of environments involves simulating stochastic agent policies, and small changes in the environment may result in large changes in the emergent agent behaviors~\citep{sturtevant2020unexpected}.

We make the following contributions: 
(1) We propose the use of deep surrogate models to predict agent performance in new environments. 
Our algorithm, Deep Surrogate Assisted Generation of Environments (\dsage{}) (\fref{fig:alg}), integrates deep surrogate models into quality diversity optimization to efficiently generate diverse environments.
(2) We show in two benchmark domains from previous work, a Maze domain~\citep{dennis2020emergent,parker2022evolving} with a trained ACCEL agent~\citep{parker2022evolving} and a Mario domain~\citep{marioai,fontaine2021illuminating} with an A* agent~\citep{baumgarten}, that \dsage{} outperforms state-of-the-art QD algorithms in discovering diverse agent behaviors. 
(3) We show with ablation studies that training the surrogate model with ancillary agent behavior data and downsampling a subset of solutions from the surrogate archive results in substantial improvements in performance, compared to the surrogate models of previous work~\citep{zhang2021deep}.

\section{Problem Definition}
\label{sec:problem}

\noindent\textbf{Quality diversity (QD) optimization.}
We adopt the QD problem definition from previous work~\citep{fontaine2021dqd}. A QD optimization problem specifies an objective function $f:\mathbb{R}^n\rightarrow\mathbb{R}$ and a joint measure function $\vm:\mathbb{R}^n\rightarrow\mathbb{R}^m$. 
For each element $s \in S$, where $S \subseteq \mathbb{R}^m$ is the range of the measure function, the QD goal is to find a solution $\vtheta \in \mathbb{R}^n$ such that $\vm(\vtheta)=s$ and $f(\vtheta)$ is maximized.

Since the range of the measure function can be continuous, we restrict ourselves to algorithms from the \mapelites{} family \citep{cully2015,mouret2015illuminating} that discretize this space into a finite number of $M$ cells. 
A solution $\vtheta$ is mapped to a cell based on its measure $\vm(\vtheta)$. 
The solutions that occupy cells form an \textit{archive} of solutions.  
Our goal is to find solutions $\vtheta_i, i \in \{1,...,M\}$ that maximize the objective $f$ for all cells in the measure space. 
\begin{equation}
  \max_{\vtheta_i} \sum_{i=1}^M f(\vtheta_i)
\label{eq:objective}
\end{equation}
The computed sum in \eref{eq:objective} is defined as the QD\=/score~\citep{pugh2016qd}, where empty cells have an objective value of 0. A second metric of the performance of a QD algorithm is coverage of the measure space, defined as the proportion of cells that are filled in by solutions: $\frac{1}{M}\sum_{i=1}^M\vone_{\vtheta_i}$.

\noindent\textbf{QD for environment generation.}  We assume a single agent acting in an environment parameterized by $\vtheta \in \mathbb{R}^n$. The environment parameters can be locations of different objects or latent variables that are passed as inputs to a generative model~\citep{goodfellow:nips14}.\footnote{For consistency with the generative model literature, we use $\mathbf{z}$ instead of $\vtheta$ when denoting latent vectors} A QD algorithm generates new solutions $\vtheta$ and evaluates them by simulating the agent on the environment parameterized by $\vtheta$. The evaluation returns an objective value $f$ and measure values $\vm$. The QD algorithm attempts to generate environments that maximize $f$ but are diverse with respect to the measures $\vm$.

\section{Background and Related Work}
\label{sec:background}

\noindent\textbf{Quality diversity (QD) optimization.} QD optimization originated in the genetic algorithm community with diversity optimization~\citep{lehman2011abandoning}, the predecessor to QD. Later work introduced objectives to diversity optimization and resulted in the first QD algorithms: Novelty Search with Local Competition~\citep{lehman2011nslc} and \mapelites{}~\citep{mouret2015illuminating, cully2015}. The QD community has grown beyond its genetic algorithm roots, with algorithms being proposed based on gradient ascent~\citep{fontaine2021dqd}, Bayesian optimization~\citep{kent2020bop}, differential evolution~\citep{choi:gecco21}, and evolution strategies~\citep{fontaine2020covariance, conti2018ns, colas2020scaling}. QD algorithms have applications in damage recovery in robotics~\citep{cully2015}, reinforcement learning~\citep{tjanaka2022approximating, nilsson2021pga, cideron2020qdrl}, and generative design~\citep{gaier2018dataefficient, hagg:gecco21}. 

Among the QD algorithms, those of particular interest to us are the model-based ones.
Current model-based~\citep{bartz2016mobsurvey, Moerland2020ModelbasedRL} QD algorithms either (1) learn a surrogate model of the objective and measure functions~\citep{gaier2018dataefficient, hagg2020designing, cazenille2019exploring}, e.g. a Gaussian process or neural network, (2) learn a generative model of the representation parameters~\citep{gaier2020blackbox, rakicevic2021poms}, or (3) draw inspiration from model-based RL~\citep{Keller2020ModelBasedQS, Lim2021DynamicsAwareQF}. In particular, Deep Surrogate Assisted \mapelites{} (\dsame{})~\citep{zhang2021deep} trains a deep surrogate model on a diverse dataset of solutions generated by \mapelites{} and then leverages the model to guide \mapelites{}. However, \dsame{} has only been applied to Hearthstone deck building, a simpler prediction problem than predicting agent behavior in generated environments.
Additionally, \dsame{} is specific to \mapelites{} only and cannot run other QD algorithms to exploit the surrogate model. 
Furthermore, \dsame{} is restricted to direct search and cannot integrate generative models to generate environments that match a provided dataset.

\noindent\textbf{Automatic environment generation.} Automatic environment generation algorithms have been proposed in a variety of fields. Methods between multiple communities often share generation techniques, but differ in how each community applies the generation algorithms.

For example, in the procedural content generation (PCG) field~\citep{shaker2016procedural}, an environment generator produces video game levels that result in player enjoyment. Since diversity of player experience and game mechanics is valued in games, many level generation systems incorporate QD optimization~\citep{gravina2019procedural, fontaine2021illuminating, earle2021illuminating, khalifa2018talakat, steckel2021illuminating, schrum2020cppn2gan, sarkar2021generating}. The procedural content generation via machine learning (PCGML)~\citep{summerville2018procedural, liu2021deep} subfield studies environment generators that incorporate machine learning techniques such as Markov Chains~\citep{snodgrass2014experiments}, probabilistic graphical models~\citep{guzdial2016game}, LSTMs~\citep{summerville2016super}, generative models~\citep{volz2018mario, giacomello:gem18, torrado:cog20, sarkar2020conditional}, and reinforcement learning~\citep{khalifa2020pcgrl, earle2021learning}. Prior work~\citep{Karavolos2021AMS} has leveraged surrogate models trained on offline data to accelerate search-based PCG~\citep{togelius2011search}.

Environment generation methods have also been proposed by the scenario generation community in robotics. Early work explored automated methods for generating road layouts, vehicle arrangements, and vehicle behaviors for testing autonomous vehicles~\citep{arnold:safecomp13,mullins:av18, abey:av19, rocklage:av17,gambi:av19,sadigh2019verifying, fremont2019scenic}. Outside of autonomous vehicles, prior work~\citep{zhou2022rocus} evaluates robot motion planning algorithms by generating environments that target specific motion planning behaviors. In human-robot interaction, QD algorithms have been applied as environment generators to find failures in shared autonomy systems~\citep{fontaine:sa_rss2021} and human-aware planners tested in the collaborative Overcooked domain~\citep{fontaine2021importance}

Environment generation can also help improve the generality of RL agents. Prior work proposes directly applying PCG level generation algorithms to improve the robustness of RL~\citep{risi2020increasing, justesen2018illuminating} or to benchmark RL agents~\citep{cobbe2020leveraging}. Paired Open-ended Trailblazer (POET)~\citep{wang2019paired, wang2020enhanced} coevolves a population of both agents and environments to discover specialized agents that solve complex tasks. POET inspired a variety of open-ended coevolution algorithms~\citep{gabor2019scenario, bossens2020qed, dharna2020co, dharna2022transfer}. Later work proposes the PAIRED~\citep{dennis2020emergent}, PLR~\citep{jiang2021prioritized,jiang2021replay}, and ACCEL~\citep{parker2022evolving} algorithms that train a single generally capable agent by maximizing the regret between a pair of agents. These methods generate environments in parallel with an agent to create an automatic training curriculum. However, the authors validate these methods on human-designed environments~\citep{kirk2021survey}. Our work proposes a method that automatically generates valid environments that reveal diverse behaviors of these more general RL agents.

\section{Deep Surrogate Assisted Generation of Environments (\dsage{})}
\label{sec:algo}

\noindent\textbf{Algorithm.} We propose the Deep Surrogate Assisted Generation of Environments (\dsage{}) algorithm for discovering environments that elicit diverse agent behaviors.
Akin to the \mapelites{} family of QD algorithms, \dsage{} maintains a \textit{ground-truth archive} where solutions are stored based on their ground-truth evaluations.
Simultaneously, \dsage{} also trains and exploits a deep surrogate model for predicting the behavior of a fixed agent in new environments.
The QD optimization occurs in three phases that take place in an outer loop: model exploitation, agent simulation, and model improvement (Fig.~\ref{fig:alg}).
\aref{alg:dsage} provides the pseudocode for the \dsage{} algorithm.

The model exploitation phase (lines~\ref{alg:inner_start}--\ref{alg:inner_end}) is an inner loop that leverages existing QD optimization algorithms and the predictions of the deep surrogate model to build an archive -- referred to as the \textit{surrogate} archive -- of solutions.
The first step of this phase is to query a list of $B$ candidate solutions through the QD algorithm's \textit{ask} method. 
These solutions are environment parameters, e.g., latent vectors of a GAN, which are passed through the environment generator, e.g., a GAN, to create an environment (line~\ref{alg:inner_env_gen}).
Next, we make predictions with the surrogate model. The surrogate model first predicts data representing the agent's behavior, e.g., the probability of occupying each discretized tile in the environment (line~\ref{alg:inner_anc}), referred to as ``ancillary agent behavior data'' ($y$). 
The predicted ancillary agent behavior data ($\hat{y}$) then guides the surrogate model's downstream prediction of the objective ($\hat{f}$) and the measure values ($\hat{\vm}$) (line~\ref{alg:inner_pred}).
Finally, the QD algorithm's \textit{tell} method adds the solution to the surrogate archive based on the predicted objective and measure values.

Note that since \dsage{} is independent of the QD algorithm, the \textit{ask} and \textit{tell} methods abstract out the QD algorithm's details.
For example, when the QD algorithm is MAP-Elites or CMA-ME, \textit{tell} adds solutions if the cell in the measure space that they belong to is empty or if the existing solution in that cell has a lower objective.
For CMA-ME, \textit{tell} also includes updating internal CMA-ES parameters.

The agent simulation phase (lines~\ref{alg:elite_sel}--\ref{alg:eval_end}) inserts a subset of solutions from the surrogate archive into the ground-truth archive.
This phase begins by selecting the subset of solutions from the surrogate archive (line~\ref{alg:elite_sel}).
The selected solutions are evaluated by generating the corresponding environment (line~\ref{alg:env_gen}) and simulating a fixed agent to obtain the true objective and measure values, as well as ancillary agent behavior data (line~\ref{alg:eval}).
Evaluation data is appended to the dataset, and solutions that improve their corresponding cell in the ground-truth archive are added to that archive (lines~\ref{alg:data_add},~\ref{alg:real_archive_add}). 

In the model improvement phase (line~\ref{alg:sm_train}), the surrogate model is trained in a self-supervised manner through the supervision provided by the agent simulations and the ancillary agent behavior data. 

The algorithm is initialized by generating random solutions and simulating the agent in the corresponding environments (lines~\ref{alg:rand_gen}-\ref{alg:init_end}).
Subsequently, every outer iteration (lines~\ref{alg:outer_start}-\ref{alg:outer_end}) consists of model exploitation followed by agent simulation and ending with model improvement.

\begin{algorithm}
\LinesNumbered
\DontPrintSemicolon
\KwIn{
$N$: Maximum number of evaluations, $n_{rand}$: Number of initial random solutions, $N_{exploit}$: Number of iterations in the model exploitation phase, $B$: Batch size for the model exploitation QD optimizer
}
\KwOut{Final version of the ground-truth archive $\mathcal{A}_{gt}$}

Initialize the ground-truth archive $\mathcal{A}_{gt}$, the dataset $\mathcal{D}$, and the deep surrogate model $sm$ \;
$\mathbf{\Theta} \leftarrow$ \textit{generate\_random\_solutions}($n_{rand}$) \label{alg:rand_gen}\;
\For{$\vtheta \in \mathbf{\Theta}$}{
    $env \leftarrow g(\vtheta)$ \label{alg:init_env_gen}\;
    $f, \vm, \bm{y} \leftarrow$ \textit{evaluate}($env$) \label{alg:init_eval}\;
    $\mathcal{D} \leftarrow \mathcal{D} \cup (\vtheta, f, \vm, \bm{y})$ \label{alg:init_data_add}\;
    $\mathcal{A}_{gt} \leftarrow$ \textit{add\_solution}($\mathcal{A}_{gt}, (\vtheta, f, \vm)$) \label{alg:init_real_archive_add}\;
} \label{alg:init_end}

$evals \leftarrow n_{rand}$ \;

\While{$evals < N$}{ \label{alg:outer_start}
    Initialize a QD optimizer $qd$ with the surrogate archive $\mathcal{A}_{surrogate}$ \label{alg:inner_start}\tikzmark{top_exploit}\tikzmark{right}\;
    \For{$itr \in \{1, 2, \ldots, N_{exploit}\}$}{
        $\mathbf{\Theta} \leftarrow$ \textit{qd.ask}($B$) \;
        \For{$\vtheta \in \mathbf{\Theta}$}{
            $env \leftarrow g(\vtheta)$ \label{alg:inner_env_gen}\;
            $\hat{y} \leftarrow$ \textit{sm.predict\_ancillary}($env$) \label{alg:inner_anc}\;
            $\hat{f}, \hat{\vm} \leftarrow$ \textit{sm.predict}($env,\hat{y}$) \label{alg:inner_pred}\;
            \textit{qd.tell}($\vtheta, \hat{f}, \hat{\vm}$) \;
        }
    }\label{alg:inner_end}
   \tikzmark{bottom_exploit}
   $\mathbf{\Theta} \leftarrow$ \textit{select\_solutions}($\mathcal{A}_{surrogate}$) \label{alg:elite_sel}\tikzmark{top_sim}\;
    \For{$\vtheta \in \mathbf{\Theta}$}{
        $env \leftarrow g(\vtheta)$ \label{alg:env_gen}\;
        $f, \vm, \bm{y} \leftarrow$ \textit{evaluate}($env$) \label{alg:eval}\;
        $\mathcal{D} \leftarrow \mathcal{D} \cup (\vtheta, f, \vm, \bm{y})$ \label{alg:data_add}\;
        $\mathcal{A}_{gt} \leftarrow$ \textit{add\_solution}($\mathcal{A}_{gt}, (\vtheta, f, \vm)$) \label{alg:real_archive_add}\;
        $evals \leftarrow evals + 1$ \tikzmark{bottom_sim} 
        \;
    } \label{alg:eval_end}
    \textit{sm.train}($\mathcal{D}$) \label{alg:sm_train} \tikzmark{top_imp}\tikzmark{bottom_imp}
    \AddNoteHacked{top_exploit}{bottom_exploit}{right}{algo_exploit}{Model Exploitation}
    \AddNoteHackedTwo{top_sim}{bottom_sim}{right}{algo_sim}{Agent Simulation}
    \AddNote{top_imp}{bottom_imp}{right}{algo_imp}{Model Improvement}
    \;
} \label{alg:outer_end}

\caption{Deep Surrogate Assisted Generation of Environments (\dsage{})}
\label{alg:dsage}
\end{algorithm}

\noindent\textbf{Self-supervised prediction of ancillary agent behavior data.} By default, a surrogate model directly predicts the objective and measure values based on the initial state of the environment and the agent (provided in the form of a one-hot encoded image).
However, we anticipate that direct prediction will be challenging in some domains, as it requires understanding the agent's trajectory in the environment.
Thus, we provide additional supervision to the surrogate model in \dsage{} via a two-stage self-supervised process.

First, a deep neural network predicts ancillary agent behavior data. In our work, we obtain this data by recording the expected number of times the agent visits each discretized tile in the environment, resulting in an ``occupancy grid.''
We then concatenate the predicted ancillary information, i.e., the predicted occupancy grid, with the one-hot encoded image of the environment and pass them through another deep neural network to obtain the predicted objective and measure values. 
We use CNNs for both predictors and include architecture details in Appendix~\ref{app:model}. 
As a baseline, we compare our model with a CNN that directly predicts the objective and measure values without the help of ancillary data.

\noindent\textbf{Downsampling to select solutions from the surrogate archive.} After the model exploitation phase, the surrogate archive is populated with solutions that were predicted to be high-performing and diverse.
Hence, a basic selection mechanism (line~\ref{alg:elite_sel}) would select all solutions from the surrogate archive, identical to \dsame{}~\citep{zhang2021deep}. 
However, if the surrogate archive is overly populated, full selection may result in a large number of ground-truth evaluations per outer-loop iteration, leading to fewer outer loops and less surrogate model training. 
To balance the trade-off between evaluating solutions from the surrogate archive and training the surrogate model, we only select a subset of solutions for evaluation by downsampling the surrogate archive. 
Downsampling uniformly divides the surrogate archive into sub-regions of cells and selects a random solution from each area. %

\section{Domains}
\label{sec:domains}

We test our algorithms in two benchmark domains from prior work: a Maze domain~\citep{gym_minigrid,dennis2020emergent,parker2022evolving} with a trained ACCEL agent~\citep{parker2022evolving} and a Mario domain~\citep{togelius20102009,fontaine2021illuminating} with an A* agent~ \citep{baumgarten}. 
We select these domains because, despite their relative simplicity (each environment is represented as a 2D grid of tiles), agents in these environments exhibit complex and diverse behaviors.

In the Maze domain, we directly search for different mazes, with the QD algorithm returning the layout of the maze. In the Mario domain, we search for latent codes that are passed through a pre-trained GAN, similar to the corresponding previous work.

We select the objective and measure functions as described below. Since the agent or the environment dynamics are stochastic in each domain, we average the objective and measure values over 50 episodes in the Maze domain and 5 episodes in the Mario domain.

\noindent\textbf{Maze.}
We set a binary objective function $f$ that is 1 if the generated environment is solvable and 0 otherwise, indicating the validity of the environment.
Since we wish to generate visually diverse levels that offer a range of difficulty level for the agent, we select as measures (1) \textit{number of wall cells} (range: $[0,256]$), and (2) \textit{mean agent path length} (range: $[0,648]$, where 648 indicates a failure to reach the goal). 

\noindent\textbf{Mario.} 
Since we wish to generate playable levels, we set the objective as the \textit{completion rate}, i.e., the proportion of the level that the agent completes before dying. 
We additionally want to generate environments that result in qualitatively different agent behaviors, thus we selected as measures: (1) \textit{sky tiles}, the number of tiles of a certain type that are in the top half of the 2D grid (range: $[0,150]$), (2) \textit{number of jumps}, the number of times that the A* agent jumps during its execution (range: $[0,100]$). 

See Appendix~\ref{app:envs} for further environment details.

\section{Experiments}
\label{sec:exp}

\subsection{Experiment Design}

\noindent\textbf{Independent variables.} In each domain (Maze and Mario), we follow a between-groups design, where the independent variable is the algorithm. We test the following algorithms:

\textit{\dsage{}}: The proposed algorithm that includes predicting ancillary agent behavior data and downsampling the surrogate archive (\sref{sec:algo}).

\textit{\dsageoa{}}: The proposed algorithm with ancillary data prediction and no downsampling, i.e., selecting all solutions from the surrogate archive.

\textit{\dsageod{}}: The proposed algorithm with downsampling and no ancillary data prediction.

\textit{\dsageb{}}: The basic version of the proposed algorithm that selects all solutions from the surrogate archive and does not predict ancillary data.

\textit{Baseline QD}: The QD algorithm without surrogate assistance. We follow previous work \citep{fontaine2021illuminating} and use CMA-ME for the Mario domain. Since CMA-ME operates only in continuous spaces, we use MAP-Elites in the discrete Maze domain. 

\textit{Domain Randomization (DR) \citep{jakobi1998,sadeghi2017,tobin2017}}: Algorithm that generates and evaluates random solutions, i.e., wall locations in the maze domain and the latent code to pass through the GAN in the Mario domain.

\noindent\textbf{Dependent variables.} 
We measure the quality and diversity of the solutions with the QD\=/score metric~\cite{pugh2016qd}(\eref{eq:objective}).
As an additional metric of diversity, we also report the archive coverage. 
We run each algorithm for 5 trials in each domain. 

\noindent\textbf{Hypothesis.}
\textit{We hypothesize that \dsage{} will result in a better QD\=/score than \dsageb{} in all domains, which in turn will result in better performance than the baseline QD algorithm. \dsage{}, \dsageb{}, and the baseline QD algorithm will all exceed DR.} 
We base this hypothesis on previous work~\citep{mouret2015illuminating, fontaine:sa_rss2021} which shows that QD algorithms outperform random sampling in a variety of domains, as well as previous work~\citep{gaier2018dataefficient,zhang2021deep} which shows that surrogate-assisted \mapelites{} outperforms standard \mapelites{} in design optimization and Hearthstone domains. 
Furthermore, we expect that the additional supervision through ancillary agent behavior data and downsampling will result in \dsage{} performing significantly better than \dsageb{}.

\begin{figure}
    \centering
    \scriptsize
    \begin{subfigure}{0.49\textwidth}
        \caption{Maze}
        \label{fig:maze-results}
        \includegraphics[width=\textwidth]{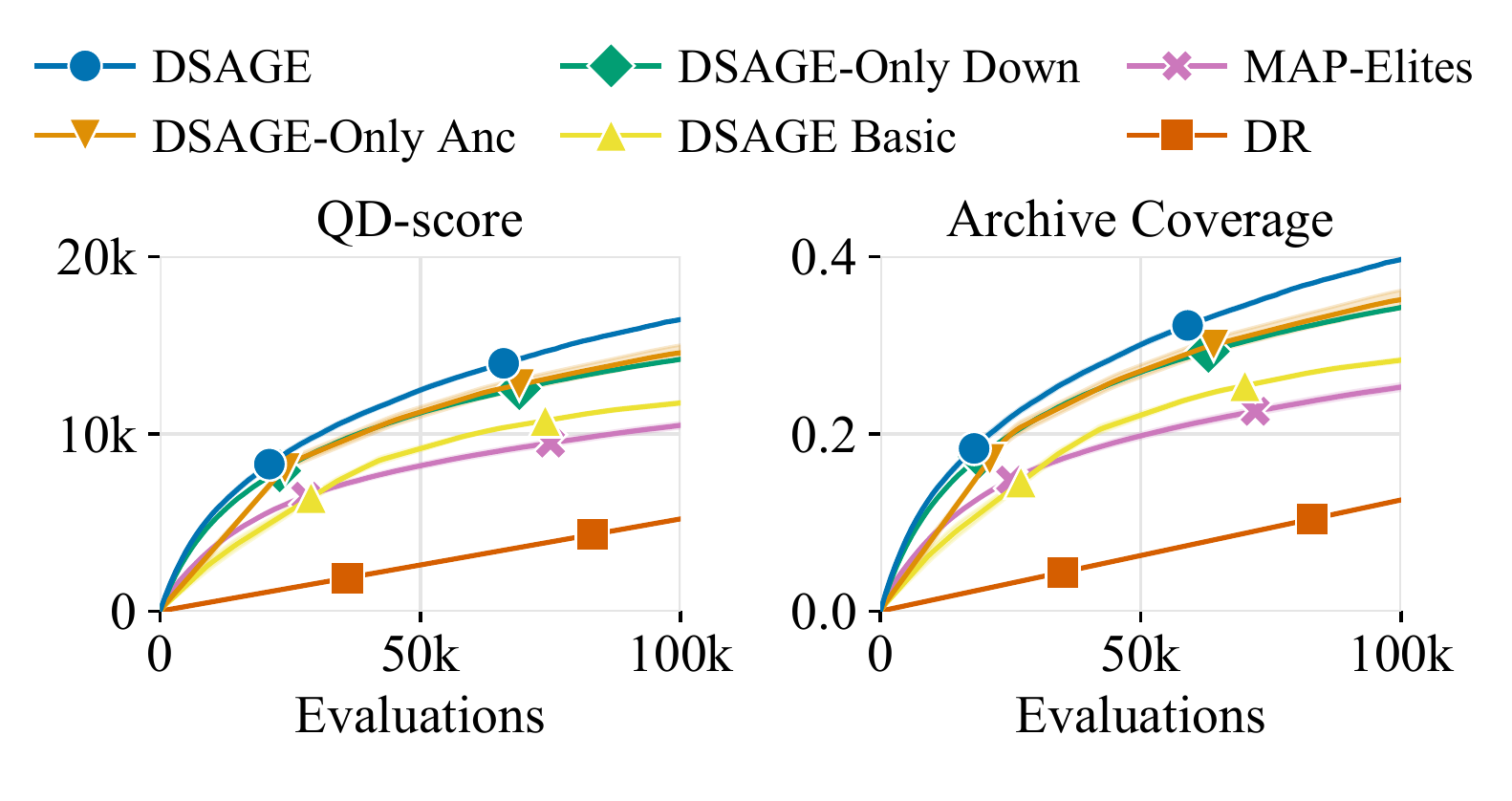}
        \begin{tabular}{L{0.85in} L{0.8in} L{0.5in}}
        \toprule
        Algorithm &                QD\=/score & Archive Coverage \\
        \midrule
        \dsage{}    &   \textbf{16,446.60 $\pm$ 42.27} &  \textbf{0.40 $\pm$ 0.00} \\
        \dsageoa{}  &  14,568.00 $\pm$ 434.56 &  0.35 $\pm$ 0.01 \\
        \dsageod{}  &   14,205.20 $\pm$ 40.86 &  0.34 $\pm$ 0.00 \\
        \dsageb{}   &   11,740.00 $\pm$ 84.13 &  0.28 $\pm$ 0.00 \\
        \mapelites{}&  10,480.80 $\pm$ 150.13 &  0.25 $\pm$ 0.00 \\
        DR & 5,199.60 $\pm$ 30.32 & 0.13 $\pm$ 0.00 \\
        \bottomrule
        \end{tabular}
    \end{subfigure}%
    \begin{subfigure}{0.49\textwidth}
        \caption{Mario}
        \label{fig:mario-results}
        \includegraphics[width=\textwidth]{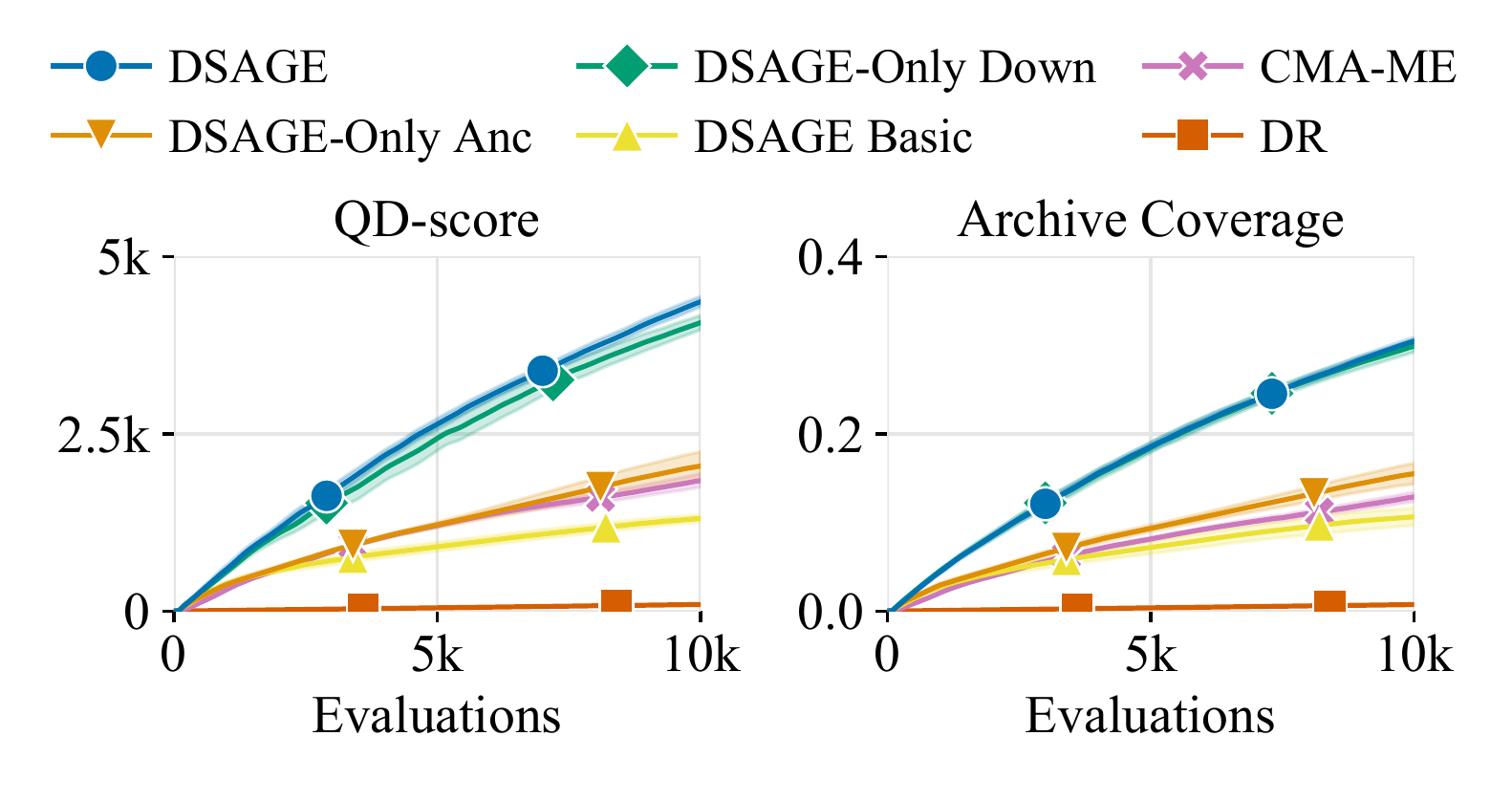}
        \begin{tabular}{L{0.85in} L{0.78in} L{0.5in}}
        \toprule
        Algorithm &               QD\=/score & Archive Coverage \\
        \midrule
        \dsage{}    &   \textbf{4,362.29 $\pm$ 72.54} &  \textbf{0.30 $\pm$ 0.00} \\
        \dsageoa{}  &  2,045.28 $\pm$ 201.64 &  0.16 $\pm$ 0.01 \\
        \dsageod{}  &  4,067.42 $\pm$ 102.06 &  \textbf{0.30 $\pm$ 0.01} \\
        \dsageb{}   &   1,306.11 $\pm$ 50.90 &  0.11 $\pm$ 0.01 \\
        \cmame{}    &   1,840.17 $\pm$ 95.76 &  0.13 $\pm$ 0.01 \\
        DR & 92.75 $\pm$ 3.01 & 0.01 $\pm$ 0.00 \\
        \bottomrule
        \end{tabular}
    \end{subfigure}
    \caption{QD\=/score and archive coverage attained by baseline QD algorithms and \dsage{} in the Maze and Mario domains over 5 trials. Tables and plots show mean and standard error of the mean.}
    \label{fig:plots}
\end{figure}

\subsection{Analysis}
\label{sec:results}

\fref{fig:plots} summarizes the results obtained by the six algorithms on the Maze and the Mario domains. 

One-way ANOVA tests showed a significant effect of the algorithm on the QD\=/score  for the Maze ($F(5, 24) = 430.98, p<0.001$) and 
Mario ($F(5, 24) = 238.09, p<0.001$) domains.

Post-hoc pairwise comparisons with Bonferroni corrections showed that \dsage{} outperformed \dsageb{}, Baseline QD, and DR in both the Maze and the Mario domains ($p<0.001$).
Additionally, \dsageb{} outperformed \mapelites{} and DR in the Maze domain ($p<0.001)$, while it performed significantly worse than the QD counterpart, \cmame{}, in the Mario domain ($p=0.003)$.
Finally, Baseline QD outperformed DR in both the Maze and Mario domains ($p<0.001$).

These results show that deep surrogate assisted generation of environments results in significant improvements compared to quality diversity algorithms without surrogate assistance. 
They also show that adding ancillary agent behavior data and downsampling is important in both domains.
Without these components, \dsageb{} has limited or no improvement compared to the QD algorithm without surrogate assistance.
Additionally, domain randomization is significantly worse than \dsage{} as well as the baselines. The archive coverage and consequently the QD\=/score is negligible in the Mario domain since randomly sampled latent codes led to little diversity in the levels.

\tref{tab:evals_for_qd} shows another metric of the speed-up provided by \dsage{}: the number of evaluations (agent simulations) required to reach a fixed QD\=/score. We set this fixed QD\=/score to be 10480.8 in the Maze domain and 1306.11 in the Mario domain, which are the mean QD\=/scores of \mapelites{} and \dsageb{}, respectively, in those domains. \dsage{} reaches these QD\=/scores faster than the baselines do.

\begin{table}
    \centering
    \caption{Number of evaluations required to reach a QD\=/score of 10480.8 in the Maze domain and 1306.11 in the Mario domain.}
    \label{tab:evals_for_qd}
    \scriptsize
    \begin{subtable}{0.49\textwidth}
        \centering
        \caption{Maze}
        \begin{tabular}{L{0.85in} L{1.1in}}
        \toprule
        Algorithm & Evaluations \\
        \midrule
        \dsage{}    &   \textbf{33,930.40 $\pm$ 1,411.04} \\
        \dsageoa{}  &  51,919.60 $\pm$ 8,254.24 \\
        \dsageod{}  &   42,816.60 $\pm$ 691.38 \\
        \dsageb{}   &   85,328.60 $\pm$ 2,947.24 \\
        \mapelites{}&  100,000 \\
        \bottomrule
        \end{tabular}
    \end{subtable}%
    \begin{subtable}{0.49\textwidth}
        \centering
        \caption{Mario}
        \begin{tabular}{L{0.85in} L{1.1in}}
        \toprule
        Algorithm & Evaluations \\
        \midrule
        \dsage{}    &   \textbf{2,464.40 $\pm$ 356.36} \\
        \dsageoa{}  &  7,727.40 $\pm$ 1,433.33 \\
        \dsageod{}  &  \textbf{2,768.60 $\pm$ 586.34} \\
        \dsageb{}   &   10,000 \\
        \cmame{}    &   5,760.00 $\pm$ 516.14 \\
        \bottomrule
        \end{tabular}
    \end{subtable}
\end{table}

To assess the quality of the trained surrogate model, we create a combined dataset consisting of data from one run of each surrogate assisted algorithm. We use this dataset to evaluate the surrogate models trained from separate runs of \dsage{} and its variants. 
\tref{tab:pred_results_main} shows the mean absolute error (MAE) of the predictions by the surrogate models.
The model learned by \dsageb{} fails to predict the agent-based measures well. It has an MAE of 157.69 for the mean agent path length in Maze and MAE = 10.71 for the number of jumps in Mario. In contrast, the model learned by \dsage{} makes more accurate predictions, with MAE = 96.58 for mean agent path length and MAE = 7.16 for number of jumps. 
We provide detailed results of the surrogate model predictions in Appendix~\ref{app:model_pred}.

\begin{table}
    \centering
    \caption{Mean absolute error of the objective and measure predictions by the surrogate models.}
    \label{tab:pred_results_main}
    \scriptsize
    \begin{tabular}{L{0.15\textwidth}C{0.07\textwidth}C{0.12\textwidth}C{0.14\textwidth}C{0.07\textwidth}C{0.11\textwidth}C{0.11\textwidth}}
        \toprule
        & \multicolumn{3}{c}{Maze} & \multicolumn{3}{c}{Mario} \\
        \cmidrule(l{2pt}r{2pt}){2-4} \cmidrule(l{2pt}r{2pt}){5-7}
        Algorithm & Objective MAE & Number of Wall Cells MAE & Mean Agent Path Length MAE & Objective MAE & Number of Sky Tiles MAE & Number of Jumps MAE \\
        \midrule
        \dsage{}    & 0.03 & 0.37 & 96.58 & 0.10 & 1.10 & 7.16  \\
        \dsageoa{}  & 0.04 & 0.96 & 95.14 & 0.20 & 1.11 & 9.97  \\
        \dsageod{}  & 0.10 & 0.95 & 151.50 & 0.11 & 0.87 & 6.52 \\
        \dsageb{}   & 0.18 & 5.48 & 157.69 & 0.20 & 2.16 & 10.71 \\
        \bottomrule
    \end{tabular}
\end{table}

\subsection{Ablation Study}
\label{sec:ablation}

\sref{sec:algo} describes two key components of \dsage{}: (1) self-supervised prediction of ancillary agent behavior data, and (2) downsampling to select solutions from the surrogate archive.
We perform an ablation study by treating the inclusion of ancillary data prediction (ancillary data / no ancillary data) and the method of selecting solutions from the surrogate archive (downsampling / full selection) as independent variables. 
A two-way ANOVA for each domain showed no significant interaction effects. 
We perform a main effects analysis for each independent variable.

\noindent\textbf{Inclusion of ancillary data prediction.}
A main effects analysis for the inclusion of ancillary data prediction showed that algorithms that predict ancillary agent behavior data (\dsage{}, \dsageoa{}) performed significantly better than their counterparts with no ancillary data prediction (\dsageod{}, \dsageb{}) in both domains ($p<0.001$).

\fref{fig:plots} shows that predicting ancillary agent behavior data also resulted in a larger mean coverage for Maze, while it has little or no improvement for Mario. 
Additionally, as shown in \tref{tab:pred_results_main}, predicting ancillary agent behavior data helped improve the prediction of the mean agent path length in the Maze domain but provided little improvement to the prediction of the number of jumps in the Mario domain.
The reason is that in the Maze domain, the mean agent path length is a scaled version of the sum of the agent's tile occupancy frequency, hence the two-stage process which predicts the occupancy grid first is essential for improving the accuracy of the model. 
On the other hand, the presence of a jump in Mario depends not only on cell occupancy, but also on the structure of the level and the sequence of the occupied cells.

\noindent\textbf{Method of selecting solutions from the surrogate archive.}  A main effects analysis for the method of selecting solutions from the surrogate archive showed that the algorithms with downsampling (\dsage{}, \dsageod{}) performed significantly better than their counterparts with no downsampling (\dsageoa{}, \dsageb{}) in both domains ($p<0.001$).

A major advantage of downsampling is that it decreases the number of ground-truth evaluations in each outer iteration. Thus, for a fixed evaluation budget, downsampling results in a greater number of outer iterations. For instance, in the Maze domain, runs without downsampling had only 6-7 outer iterations, while runs with downsampling had approximately 220 outer iterations. 
More outer iterations leads to more training and thus higher accuracy of the surrogate model. 
In turn, a more accurate surrogate model will generate a better surrogate archive in the inner loop. 

We include an ablation in Appendix~\ref{app:longer_training} to test between two possible explanations for why having more outer iterations helps with performance: (1) larger number of training epochs, (2) more updates to the dataset allowing the surrogate model to iteratively correct its own errors. We observed that iterative correction accounted for most of the performance increase with downsampling.

The second advantage of downsampling is that it selects solutions evenly from all regions of the measure space, thus creating a more balanced dataset. This helps train the surrogate model in parts of the measure space that are not frequently visited. We include an additional baseline in Appendix~\ref{app:random_sample} in which we select a subset of solutions uniformly at random from the surrogate archive instead of downsampling.
We observe that downsampling has a slight advantage over uniform random sampling in the Maze domain.

Furthermore, if instead of downsampling we sampled multiple solutions from nearby regions of the surrogate archive, the prediction errors could cause the solutions to collapse to a single cell in the ground-truth archive, resulting in many solutions being discarded.

Overall, our ablation study shows that both predicting the occupancy grid as ancillary data and downsampling the surrogate archive independently help improve the performance of \dsage{}. 

\subsection{Qualitative Results}
\begin{figure}
    \centering
    \includegraphics[width=\linewidth]{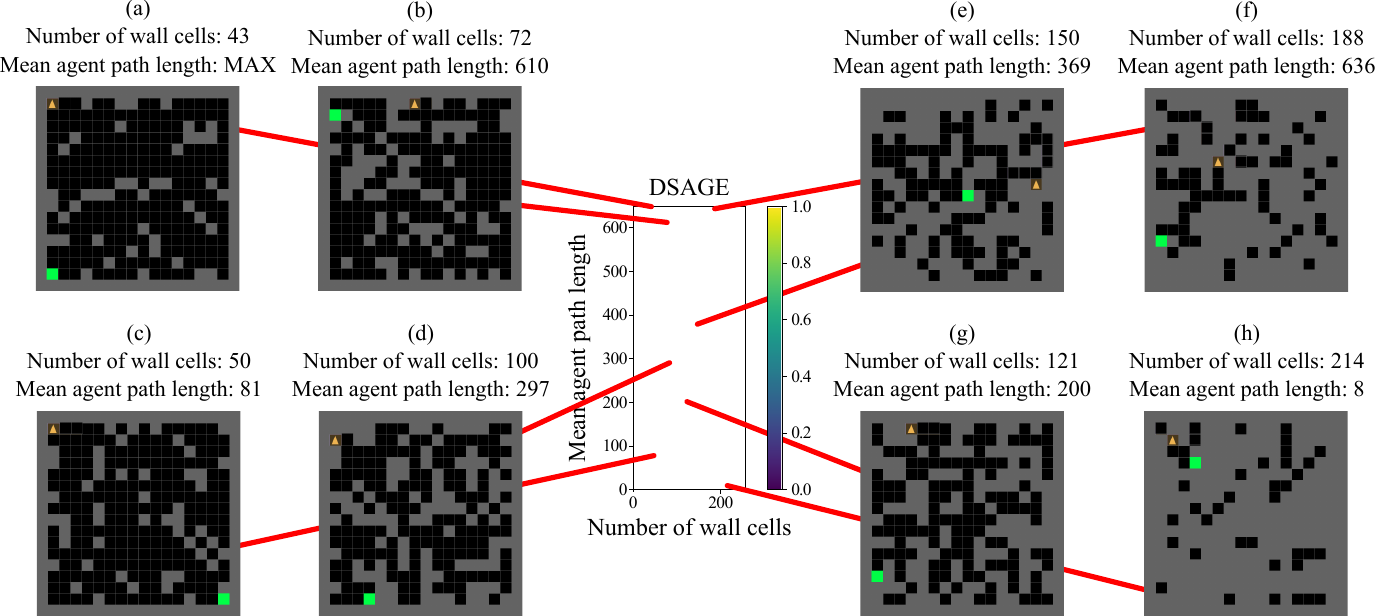}
    \caption{Archive and levels generated by \dsage{} in the Maze domain. The agent's initial position is shown as an orange triangle, while the goal is a green square.}
    \label{fig:maze_heatmap}
\end{figure}

\begin{figure}
    \centering
    \includegraphics[width=\linewidth]{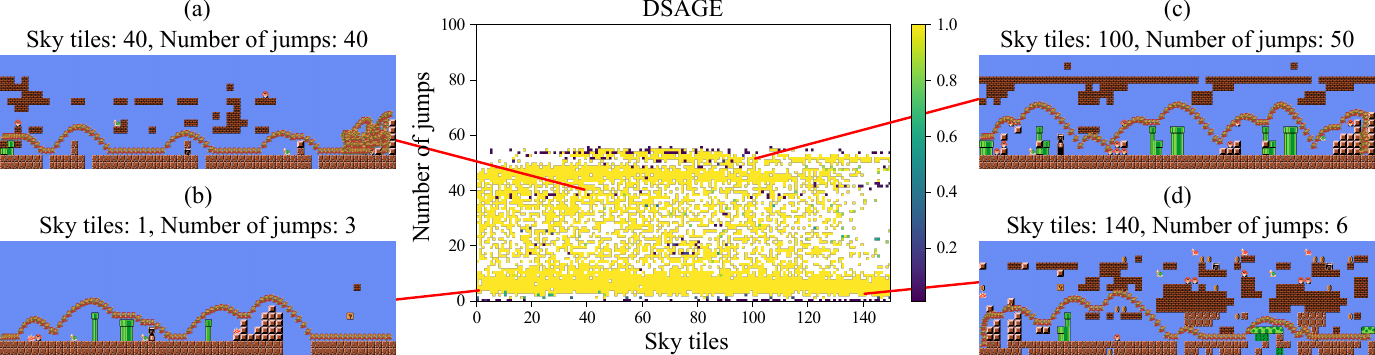}
    \caption{Archive and levels generated by \dsage{} in the Mario domain. Each level shows the path Mario takes, starting on the left of the level and finishing on the right.}
    \label{fig:mario_heatmap}
\end{figure}

\fref{fig:maze_heatmap} and \fref{fig:mario_heatmap} show example environments generated by \dsage{} in the Maze and Mario domains. 

Having the mean agent path length as a measure in the Maze domain results in environments of varying difficulty for the ACCEL agent. 
For instance, we observe that the environment in \fref{fig:maze_heatmap}(a) has very few walls, yet the ACCEL agent gets stuck in the top half of the maze and is unable to find the goal within the allotted time. On the other hand, the environment in \fref{fig:maze_heatmap}(d) is cluttered and there are multiple dead-ends, yet the ACCEL agent is able to reach the goal. 

\fref{fig:mario_heatmap} shows that the generated environments result in qualitatively diverse behaviors for the Mario agent too. Level (b) only has a few sky tiles and is mostly flat, resulting in a small number of jumps. Level (c) has a ``staircase trap'' on the right side, forcing the agent to perform continuous jumps to escape and complete the level. We include videos of the playthroughs in the supplemental material. 

\section{Societal Impacts}
\label{sec:impact}

By introducing surrogate models into quality diversity algorithms, we can efficiently generate environments that result in diverse agent behaviors. While we focused on an RL agent in a Maze domain and a symbolic agent in a Mario game domain, our method can be applied to a variety of agents and domains. This can help with testing the robustness of agents, attaining insights about their behavior, and discovering edge cases before real-world deployment. Furthermore, we anticipate that in the future, closing the loop between environment generation and agent training can improve the ability of agents to generalize to new settings and thus increase their widespread use. 

Our work may also have negative impacts. Training agents in diverse environments can be considered as a step towards open-ended evolution~\citep{stanley2017open}, which raises concerns about the predictability and safety of the emergent agent behaviors~\citep{ecoffet2020open,hendrycks2021unsolved}. Discovering corner cases that result in unwanted behaviors or catastrophic failures may also be used maliciously to reveal vulnerabilities in deployed agents~\citep{roy2018evolutionary}.

\section{Limitations and Future Work}
\label{sec:limitations}

Automatic environment generation is a rapidly growing research area with a wide range of applications, including designing video game levels~\citep{shaker2016procedural,summerville2018procedural,liu2021deep}, training and testing autonomous agents~\citep{risi2020increasing,cobbe2020leveraging,wang2019paired,dennis2020emergent,parker2022evolving}, and discovering failures in human-robot interaction~\citep{fontaine:sa_rss2021,fontaine2021illuminating}. We introduce the \dsage{} algorithm, which efficiently generates a diverse collection of environments via deep surrogate models of agent behavior. 

Our paper has several limitations. First, occupancy grid prediction does not encode temporal information about the agent. While this prediction allows us to avoid the compounding error problem of model-based RL~\citep{xiao2019learning}, forgoing temporal information makes it harder to predict some behaviors, such as the number of jumps in Mario.  We will explore this trade-off in future work. 
 
Furthermore, we have studied 2D domains where a single ground-truth evaluation lasts between a few seconds and a few minutes. We are excited about the use of surrogate models to predict the performance of agents in more complex domains with expensive, high-fidelity simulations~\citep{wurman2022outracing}.

\begin{ack}

This work was partially funded by NSF CAREER (\#2145077) and NSF GRFP (\#DGE-1842487). One of the GPUs used in the experiments was awarded by the NVIDIA Academic Hardware Grant. We thank J. Parker-Holder et al., the authors of the ACCEL agent, for providing a pre-trained model of the agent for our experiments. We also thank Ya-Chuan Hsu for providing invaluable feedback.
\end{ack}

\begin{small}
\bibliographystyle{ieeetr}
\bibliography{ref}
\end{small}

\newpage 

\appendix

\section{Environment Details}
\label{app:envs}

\subsection{Maze}

\paragraph{Environment.}

The mazes that we consider in this paper are implemented as MiniGrid environments~\citep{gym_minigrid}.
Each maze is a $16 \times 16$ grid containing walls and empty cells, two of which are the starting and the goal cells. 
An agent solving the maze starts at the starting cell and observes a $5 \times 5$ area around itself. 
The agent can move forward into an empty cell or turn left or right in its own cell.
To maintain consistency with the MiniGrid environments, the agent is also allowed to pick up, drop or toggle an object or notify that a task is done.
In the mazes generated by our work, all those actions result in the agent staying in the same cell.
A time limit of $648$ is used since an optimal agent will be able to finish all possible $16 \times 16$ mazes in this duration.
If the agent reaches the goal within this time limit, it receives a reward of $1 - 0.9 \times$ fraction of the time limit used. 
Otherwise, the agent receives no reward. 

\paragraph{Environment generator.}

The environment generator accepts a $16 \times 16$ bit map denoting the walls and empty spaces as the input.
For better visualization, we add a wall surrounding the $16 \times 16$ region.
We set the starting cell and goal cell to be the pair of empty cells that are furthest apart, as identified by the Floyd-Warshall algorithm~\citep{cormen2022introduction}.

\paragraph{Agent.}

We select an agent from a recent work on open-ended learning, ACCEL \citep{parker2022evolving}, for the purpose of evaluation. 
Since individual ACCEL agents have a high variance in their performance, we evaluated the agents trained with four different random seeds on three of the test mazes given in the original paper (\textit{Labyrinth}, \textit{16Rooms}, \textit{LargeCorridor}). 
We chose the best performing agent out of the four and fixed it for all our experiments. 
The selected agent was able to always reach the goal in those test mazes.

\subsection{Mario}

\paragraph{Environment}
The Mario environments that we consider in this paper are implemented in the Mario AI Framework~\citep{marioframework,marioai}.
Each level is a $16\times56$ grid of tiles, where each tile can be one of 17 different objects.
The agent in each environment receives as input the current game state, consisting of all tiles that are visible on the screen.
The agent then outputs the action for Mario to take.
Each episode runs for 20 time ticks.

\begin{figure}[h]
    \centering
    \includegraphics[width=\linewidth]{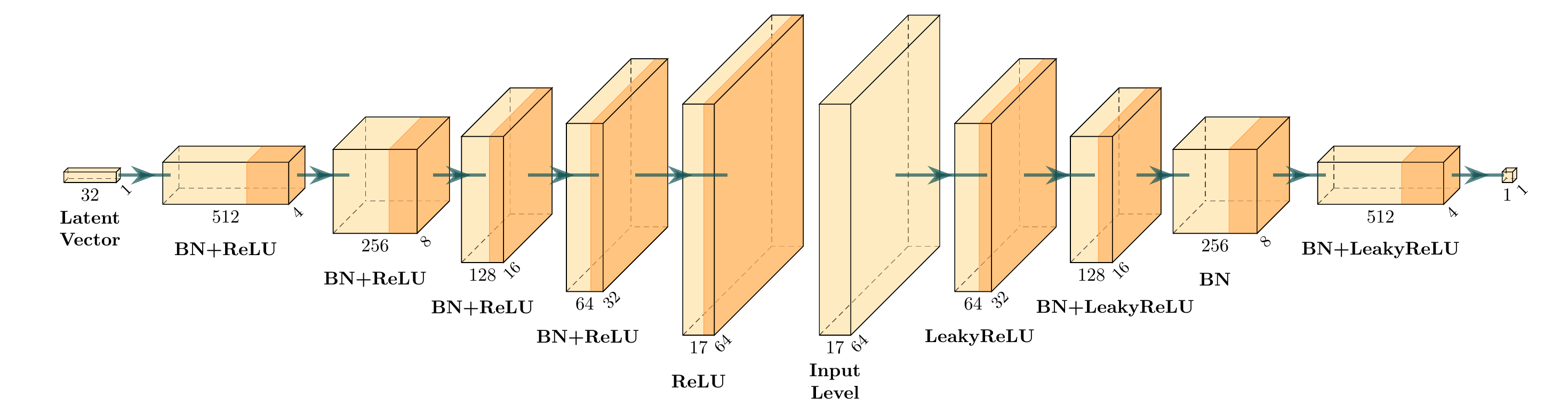}
    \caption{GAN architecture for generating Mario levels (BN stands for Batch Normalization \citep{ioffe2015batch}).}
    \label{fig:mariogan}
\end{figure}

\paragraph{Environment generator.}
Drawing from prior work \citep{fontaine2021illuminating,volz2018mario}, the Mario environments are generated with a GAN pre-trained on human-authored levels with the WGAN algorithm \citep{arjovsky2017wgan,gulrajani2017wgan}.
The GAN's generator takes as input a latent vector of size 32 and outputs a $16 \times 56$ level padded to $64 \times 64$.
The GAN architecture is shown in \fref{fig:mariogan}.

\paragraph{Agent.}
In each environment, we run the A* agent developed by Robin Baumgarten \citep{baumgarten}.
This agent won the Mario AI competitions at the ICE-GIC conference and the IEEE Computational Intelligence in Games symposium in 2009.
The trajectory taken by the agent in a level is stochastic due to randomness in the environment dynamics.

\section{Deep Surrogate Model}
\label{app:model}

\begin{figure}
  \centering
  \includegraphics[width=0.9\textwidth]{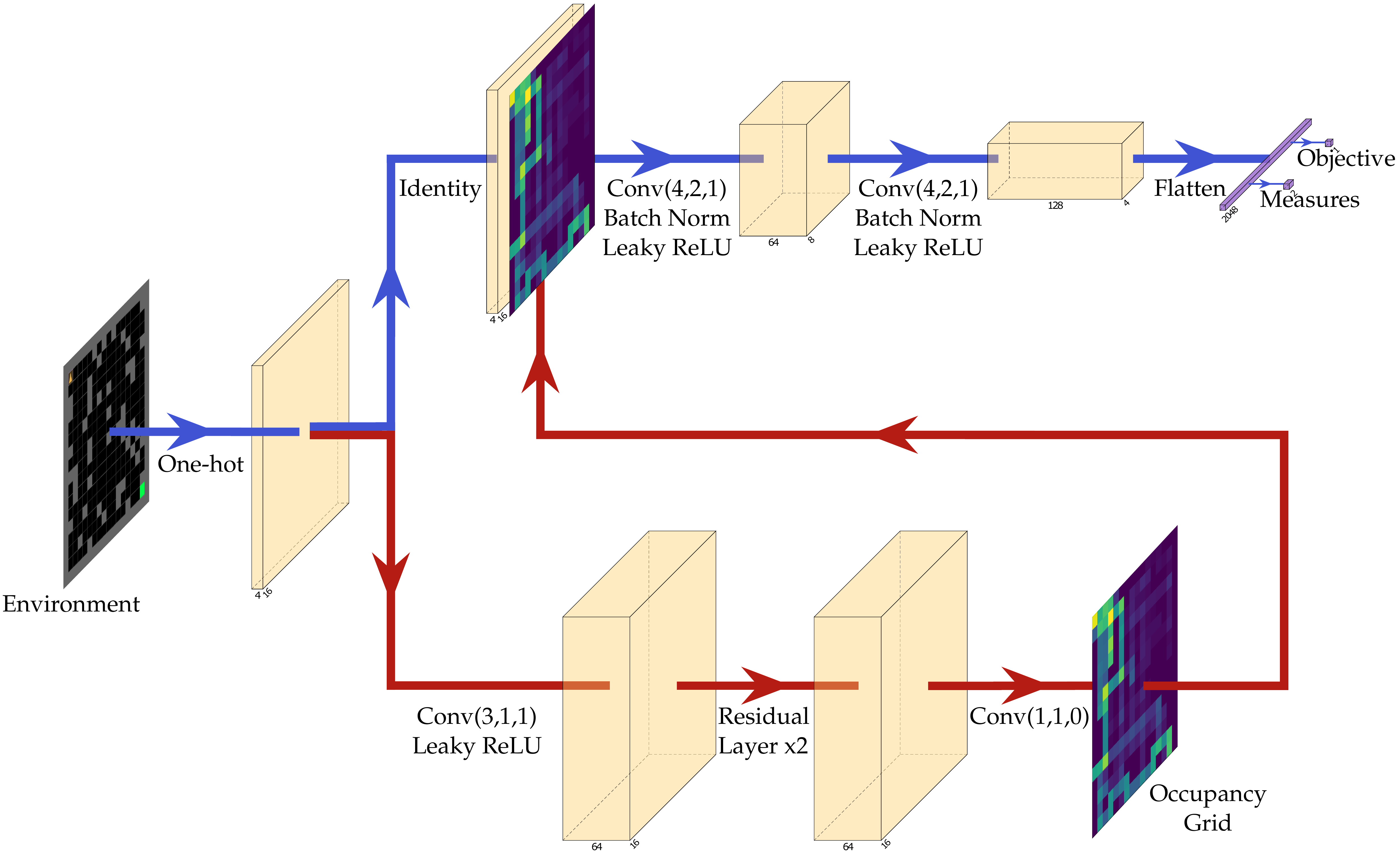}
  \caption{Architecture of the surrogate model. The model predicts the occupancy grid (red arrows) which guides the downstream prediction of the objective and the measure values (blue arrows).}
  \label{fig:cnn}
\end{figure}

In the \dsage{} algorithm, we maintain a deep surrogate model (\fref{fig:cnn}) for predicting the objective and the measures resulting from simulating an agent's execution in the environment. The input to this model, provided as a one-hot encoded representation of the image of the environment, is passed through a two-stage deep surrogate model as described in \sref{sec:algo}.

The first stage predicts the ancillary agent behavior data that is in the form of an occupancy grid.
The predictor consists of a $3 \times 3$ convolution (with Leaky ReLU activation) followed by two residual layers~\citep{he2016deep} and a $1 \times 1$ convolution.
Since the occupancy grid depends on the layout of the environment, we believe that residual layers' propagation of the input information is helpful for prediction.

The predicted occupancy grid and the one-hot encoded image of the environment are stacked and passed through another CNN that predicts the objective and the measure values. 
The architecture of this CNN is inspired by the discriminator architecture in prior work on generating Mario levels with a GAN~\citep{fontaine2021illuminating,volz2018mario}. 
The input is passed through layers of $4 \times 4$ strided convolutions with a stride of 2 and an increasing number of channels. Each convolution is followed by Batch Normalization~\citep{ioffe2015batch} and LeakyReLU activation. 
Once the height and width of the output of a convolution have been reduced to 4, it is flattened and passed through two fully connected layers to obtain the objective and the measure values. 

\dsageb{} and \dsageod{} do not predict the occupancy grid. The surrogate model in those algorithms directly predicts the objective and the measure values as denoted by the blue arrows.

\subsection{Evaluating the Prediction Performance}
\label{app:model_pred}

\noindent\textbf{Mean absolute error.} To test the prediction performance of the deep surrogate model trained by \dsage{} and its variants, we select two separate runs of each algorithm.
The datasets generated in the first run of each algorithm are combined into a single dataset.
We then evaluate the trained surrogate models from the second run of each algorithm on the combined dataset by calculating the mean absolute error (MAE) between the predicted and the true objective and measures corresponding to the solutions in the combined dataset.

\tref{tab:pred_results_main} shows the obtained MAEs in the Maze and the Mario domains.
In both domains, we observe that the measures that depend on agent behavior (mean agent path length for Maze and number of jumps for Mario) are harder to predict compared to the ones that only depend on the environment (number of wall cells for Maze and number of sky tiles for Mario).
Indeed, the MAEs for the number of wall cells in the Maze domain and the number of sky tiles in the Mario domain are much smaller than the MAE for the mean agent path length and the number of jumps, respectively.

In the Maze domain, predicting ancillary agent behavior data helped improve the prediction of the mean agent path length. 
Both \dsage{} and \dsageoa{} have better predictions compared to their counterparts that do not predict ancillary data.
Since the mean agent path length is a scaled version of the sum of the occupancy grid, having a good prediction of the occupancy grid makes the downstream prediction task much easier. 
We believe that the additional supervision during training in the form of the occupancy grid guides the surrogate model towards understanding the layout of the maze and the agent's behavior.

On the other hand, we see little improvement when predicting the number of jumps in the Mario domain.
Here, downsampling provided a larger boost to the predictions, with \dsage{} and \dsageod{} making better predictions than their counterparts without downsampling.
Since we do not store temporal information in the occupancy grid, predicting the number of jumps remains a challenging task even with an accurate prediction of the occupancy grid.
We conjecture that the increased number of outer iterations when downsampling played a more important role in correcting the errors of the surrogate model and improving its predictions.

\begin{figure}
    \centering
    \includegraphics[width=\textwidth]{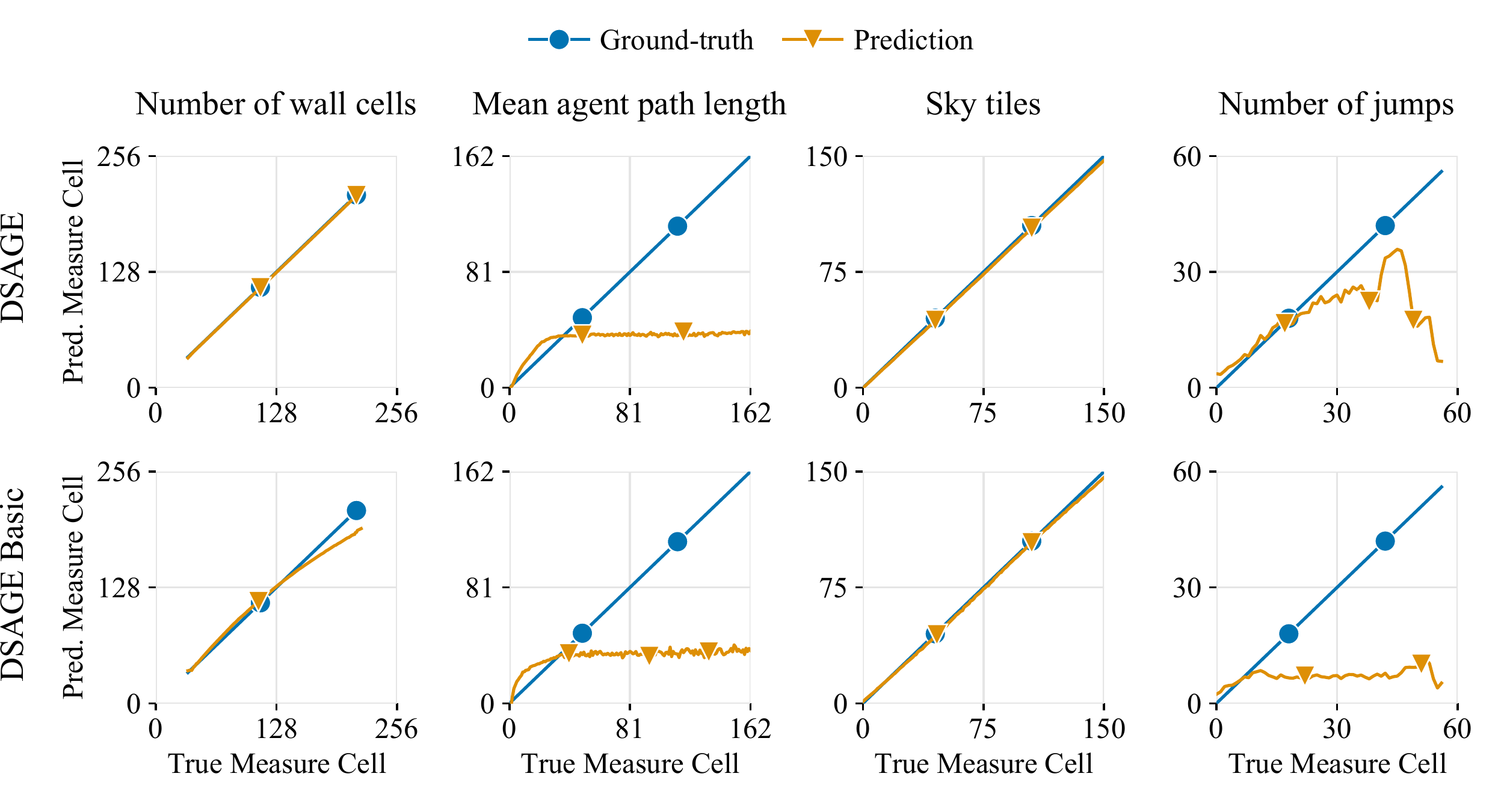}
    \caption{Correlation between the predicted and the true measure cells. The first row corresponds to \dsage{} while the second row corresponds to \dsageb{}. The columns correspond to the two measures in the Maze and the Mario domains respectively. We observe that long agent path lengths and high numbers of jumps are more difficult to predict.}
    \label{fig:preds}
\end{figure}

\noindent\textbf{Correlation plots.} To further test if \dsage{}'s predictions of some measures were more accurate in certain regions of the archive, for each solution we plot the true measure cell on the x-axis and the average of the corresponding predicted measure cell on the y-axis (\fref{fig:preds}). 
In this plot, accurate predictions would fall on the $x=y$ line (denoted in blue), and inaccurate ones would be above or below the line.

Once again, we see that the measures dependent on agent simulation, i.e., the mean agent path length in Maze and the number of jumps in Mario, are difficult to predict. 
Interestingly, we observe that accurately predicting large number of jumps and long agent path length is harder compared to predicting them when the true value is low. 
Since the agent would be revisiting the tiles multiple times when the path length or the number of jumps is high, it becomes harder to obtain useful information from the occupancy grid. 

We also believe that in these regions, minor environment differences could cause a large change in the measure value, making the prediction problem extremely difficult. 
For example, if a jump in Mario is barely possible, the agent might need to try multiple times. 
But if one block is removed to make the jump easier, the agent might be able to finish it in one try, drastically reducing the total number of jumps.

\begin{figure}
    \centering
    \includegraphics[width=\textwidth]{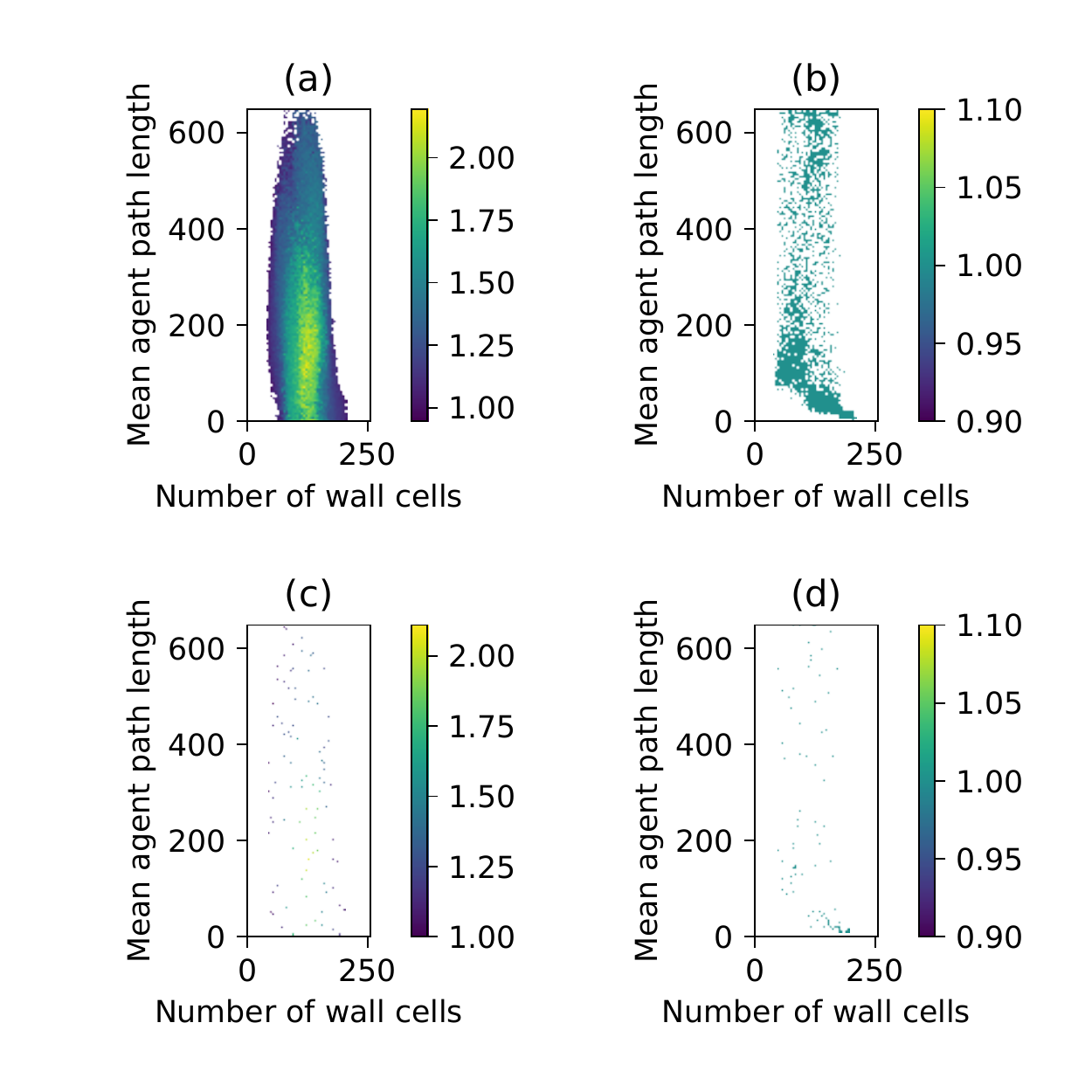}
    \caption{After running one surrogate model exploitation inner loop, we visualize the (a) surrogate archive, (b) positions of solutions from the surrogate archive in the ground-truth archive, (c) downsampled surrogate archive, and (d) positions of solutions from the downsampled surrogate archive in the ground-truth archive.}
    \label{fig:archs}
\end{figure}

\noindent\textbf{Surrogate archive accuracy.} 
To understand how the surrogate model's accuracy affects the creation of the ground-truth archive, we run an additional surrogate model exploitation inner loop starting from a completed \dsage{} run and obtain a surrogate archive. 
We evaluate all the solutions in the surrogate archive and add them to a separate archive based on the ground-truth objective and measures. 
Additionally, we downsample the surrogate archive and create a corresponding ground-truth archive from the selected solutions.

\fref{fig:archs} shows the full (\ref{fig:archs}.a) and the downsampled (\ref{fig:archs}.c) surrogate archive and the corresponding ground\=/truth archives (\ref{fig:archs}.b,~\ref{fig:archs}.d) in the Maze domain.
We observe that many of the solutions from the surrogate archive end up in the same cell in the ground-truth archive, creating holes in the ground-truth archive. 
Only 47\% and 41\% of the solutions from the surrogate archive ended up in unique cells in the Maze and the Mario domains respectively.
On the other hand, when downsampling, the percentage of surrogate archive solutions filling unique cells in the ground-truth archive improved to 97\% and 94\% in the Maze and the Mario domains respectively.
Hence, downsampling reduces the number of unnecessary ground-truth evaluations.

In the Maze domain, only 0.06\% of the surrogate archive solutions ended up in the exact same cell of the ground-truth archive as predicted. 
4.6\% of the solutions were in the $8 \times 6$ (the area from which downsampled solutions are chosen) neighborhood of the predicted cell.
The average Manhattan distance between the predicted cell and the true cell was 53.8.
In the Mario domain, 2.0\% of the solutions were exactly in the same cell, 23.3\% in the $5 \times 5$ neighborhood, and the average Manhattan distance was 14.2.

Despite the low accuracy of the surrogate model in terms of predicting the exact cell of the archive that the solution belongs to, the predictions were in the nearby region of the archive as evidenced by the average Manhattan distance.
Furthermore, we conjecture that the holes in the ground-truth archive from a single outer iteration (as seen in \fref{fig:archs}.b,~\ref{fig:archs}.d) are filled by solutions from other outer iterations.
Hence, the final ground-truth archive (\fref{fig:maze_heatmap},~\fref{fig:mario_heatmap}) is more densely filled, leading to a better archive coverage and a better QD\=/score.

\section{Experimental Details}
\label{app:exp}

\paragraph{QD Optimization Algorithm.}

In the Maze domain, we used the \mapelites{} algorithm to generate the wall and the empty tiles of a maze.
The first 100 solutions were generated by setting each cell to be either a wall cell or an empty cell uniformly at random.
Every subsequent solution was generated by first choosing a random solution in the archive and mutating 10 random cells to a random value.
The batch size was set to 150, i.e., 150 solutions were generated and evaluated in each iteration of the \mapelites{} algorithm.
The archive was divided into $256 \times 162$ cells corresponding to the number of wall cells and the mean agent path length respectively.

In the Mario domain, we followed previous work~\citep{fontaine2021illuminating} and selected the \cmame{} algorithm for QD optimization.
The archive was divided into $150 \times 100$ cells corresponding to the number of sky tiles and the number of jumps respectively.
The solutions, which are the input to a pre-trained GAN from previous work~\citep{fontaine2021illuminating}, were generated by 5 improvement emitters, each with a batch size of 30 and mutation power of 0.2.

In the baselines without a surrogate model, we ran the QD optimization algorithm until the number of ground-truth evaluations reached the given budget.
For the other algorithms, we used the QD optimizer in the surrogate model exploitation phase and ran 10,000 iterations of the corresponding algorithm to create the surrogate archive.

We implemented all QD algorithms in Python with the pyribs~\citep{pyribs} library. 

\paragraph{Ancillary data and downsampling.}

In both domains, we recorded and stored the average number of visits by the agent to each discretized tile in the environment as the ancillary data. 
Algorithms using downsampling chose a single random elite from every $8 \times 6$ cells in the Maze domain and every $5 \times 5$ cells in the Mario domain

\paragraph{Surrogate Model Training.}

At the start of each outer iteration, the deep surrogate model was trained on the most recent 20,000 data samples for 200 epochs with a batch size of 64. 
The surrogate model was updated by backpropagating the mean square error loss between the predicted and the true objective, measures, and ancillary data. 
The model weights were then updated by the Adam~\citep{adam} optimizer with a learning rate of 0.001 and betas equal to 0.9 and 0.999 respectively.
We implemented the surrogate model with the PyTorch~\citep{pytorch} library.

\paragraph{Computational Resources.}

For each algorithm-domain pair, we repeated the experiments 5 times and compared the mean performance. 
Experiments were run on two local machines and a high-performance cluster. 
The local machines had AMD Ryzen Threadripper with a 64\=/core (128 threads) CPU and an NVIDIA GeForce RTX~3090/RTX~A6000 GPU. 
16 CPU cores and one V100 GPU were allocated for each run on the cluster. 
Maze experiments without downsampling lasted for 4-5 hours while those with downsampling lasted for around 30 hours.
Mario experiments without downsampling took 2-3 hours while those with downsampling took around 12 hours.

A single ground-truth evaluation in the Maze domain took between 1 to 13 seconds, with a mean of 3.5 seconds. 
The variation was mostly due to the difference in the agent performance since mazes that were finished in fewer steps required fewer forward passes through the agent's policy network.
Evaluations in the Mario domain took between 1 to 135 seconds, with an average of 53 seconds, depending on the generated level.
In contrast, a complete inner loop involving the surrogate model exploitation phase (around 1,500,000 surrogate evaluations) finished in around 90 seconds.

\section{Ablation: Effect of More Outer Iterations}
\label{app:longer_training}

We perform an ablation to test between two possible explanations for why having more outer iterations helps with performance: One explanation is that the larger number of training epochs, resulting from training the model in each outer iteration, itself helps with the accuracy of the surrogate model~\citep{power2022grokking}. 
The second explanation is based on the fact that at the beginning of training, the surrogate model is inaccurate, and hence, the data generated by evaluating solutions in the surrogate archive would have been incorrectly predicted by the surrogate model. 
A larger number of outer iterations results in a larger number of times the algorithm updates the dataset with these adversarial examples, allowing the surrogate model to iteratively correct its own errors.

To disambiguate the two explanations, we increased the number of training epochs for the algorithms that do not use downsampling (\dsageoa{} and \dsageb{}), making the total number of training epochs the same as that with downsampling.
In the Maze domain, the surrogate models of \dsageoa{} and \dsageb{} were trained for 5300 and 6400 epochs respectively in each outer iteration, compared to 200 epochs with downsampling.
In the Mario domain, the models of \dsageoa{} and \dsageb{} were trained for 1350 epochs in each outer iteration, compared to 200 epochs with downsampling.

\begin{table}
    \centering
    \caption{QD\=/score and archive coverage attained by \dsage{} variants with original hyperparameters and longer training versions of \dsageb{} and \dsageoa{} in the Maze and Mario domains over 5 trials.}
    \label{tab:long_training_res}
    \scriptsize
    \begin{tabular}{L{0.29\textwidth}C{0.15\textwidth}C{0.13\textwidth}C{0.15\textwidth}C{0.13\textwidth}}
        \toprule
        & \multicolumn{2}{c}{Maze} & \multicolumn{2}{c}{Mario} \\
        \cmidrule(l{2pt}r{2pt}){2-3} \cmidrule(l{2pt}r{2pt}){4-5}
        Algorithm & QD\=/score & Archive Coverage & QD\=/score & Archive Coverage \\
        \midrule
        \dsage{} & 16,446.60 $\pm$ 42.27 & 0.40 $\pm$ 0.00 & 4,362.29 $\pm$ 72.54 & 0.30 $\pm$ 0.00 \\
        \dsageoa{} (longer training) & 14,936.40 $\pm$ 400.45 & 0.36 $\pm$ 0.01 & 1,679.55 $\pm$ 213.21 & 0.11 $\pm$ 0.01 \\
        \dsageoa{} & 14,568.00 $\pm$ 434.56 & 0.35 $\pm$ 0.01 & 2,045.28 $\pm$ 201.64 & 0.16 $\pm$ 0.01 \\
        \dsageod{} & 14,205.20 $\pm$ 40.86 & 0.34 $\pm$ 0.00 & 4,067.42 $\pm$ 102.06 & 0.30 $\pm$ 0.01 \\
        \dsageb{} (longer training) & 12,618.20 $\pm$ 58.94 & 0.30 $\pm$ 0.00 & 1,983.84 $\pm$ 434.15 & 0.13 $\pm$ 0.03 \\
        \dsageb{} & 11,740.00 $\pm$ 84.13 & 0.28 $\pm$ 0.00 & 1,306.11 $\pm$ 50.90 & 0.11 $\pm$ 0.01 \\
        \bottomrule
    \end{tabular}
\end{table}

\tref{tab:long_training_res} shows the results with the longer training versions of \dsageb{} and \dsageoa{}.
Longer training improves the QD\=/score and the archive coverage for both \dsageb{} and \dsageoa{} in the Maze domain and for \dsageb{} in the Mario domain, but they still perform much worse than their counterparts with downsampling, \dsageod{} and \dsage{}.
Hence, more iterative corrections of the errors of the surrogate model in variants with downsampling (due to a larger number of outer iterations) seems to be the major cause of performance improvement.

\section{Ablation: Random Selection of Surrogate Archive Solutions}
\label{app:random_sample}

As discussed in \sref{sec:ablation}, selecting solutions from the surrogate archive with downsampling has several advantages which lead to better performance, with the major advantage being that downsampling increases the number of outer loop iterations.
However, we could also increase the number of outer iterations by choosing a different subset selection mechanism, including simply selecting solutions uniformly at random. 
Thus, we test \dsage{} with the random selection mechanism as an additional baseline.
Namely, after every inner loop, we select a fixed number of solutions from the surrogate archive uniformly at random such that the number of outer iterations is approximately the same for both downsampling and random sampling.

\begin{table}
    \centering
    \caption{Mean and standard error of the QD\=/score and archive coverage attained by \dsage{} and \dsage{} with random sampling in the Maze and Mario environments over 5 trials.}
    \label{tab:rsample_results}
    \small
    \begin{tabular}{L{0.25\textwidth}C{0.17\textwidth}C{0.13\textwidth}C{0.17\textwidth}C{0.13\textwidth}}
        \toprule
        & \multicolumn{2}{c}{Maze} & \multicolumn{2}{c}{Mario} \\
        \cmidrule(l{2pt}r{2pt}){2-3} \cmidrule(l{2pt}r{2pt}){4-5}
        Algorithm & QD\=/score & Archive Coverage & QD\=/score & Archive Coverage \\
        \midrule
        \dsage{}                   & 16,446.60 $\pm$ 42.27 & 0.40 $\pm$ 0.00 & 4,362.29 $\pm$  ~~72.54 & 0.30 $\pm$ 0.00 \\
        \dsage{} (random sampling) & 15,974.40 $\pm$ 78.71 & 0.39 $\pm$ 0.00 & 4,370.28 $\pm$ 107.87 & 0.30 $\pm$ 0.01 \\
        \bottomrule
    \end{tabular}
\end{table}

\tref{tab:rsample_results} shows the results obtained by \dsage{} with downsampling and random sampling.
We observe that the performance with random sampling is lower than that of downsampling in the Maze domain, but they are very close in the Mario domain.
Hence, we can conclude that increasing the number of outer iterations is the largest contributor to the performance improvement, although downsampling has additional advantages that improve its performance in the Maze domain.

\section{Qualitative Analysis of the Algorithms}
\label{app:archives}

\begin{figure}[h]
    \centering
    \includegraphics[width=\linewidth]{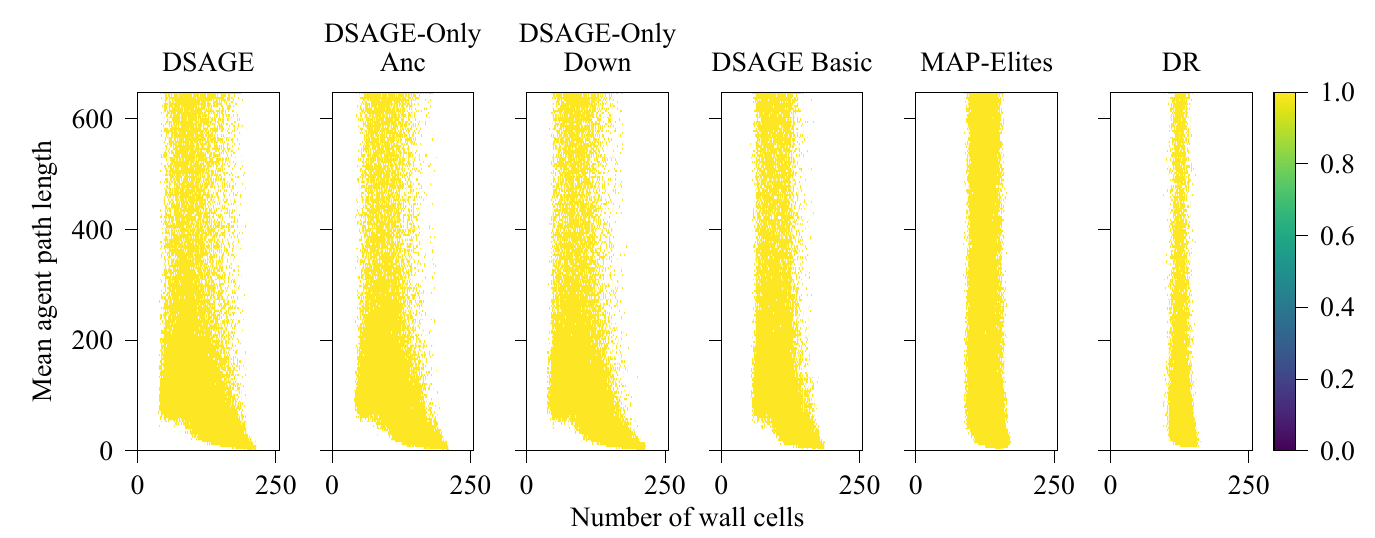}
    \caption{Example archives generated by algorithms in the Maze domain. Among the algorithms, \dsage{} fills the largest portion of the archive, resulting in the highest QD\=/score, while \mapelites{} fills the smallest portion, resulting in the lowest QD\=/score. Since the objective only tests whether the level is valid, all levels in each archive have an objective of 1. Note that certain portions of the archive are physically impossible to obtain. For example, a maze with 0 wall cells would have the starting position and the goal at opposite corners, meaning that the mean agent path length must be at least 32.}
    \label{fig:heatmap_figure_maze}
\end{figure}

\begin{figure}[h]
    \centering
    \includegraphics[width=\linewidth]{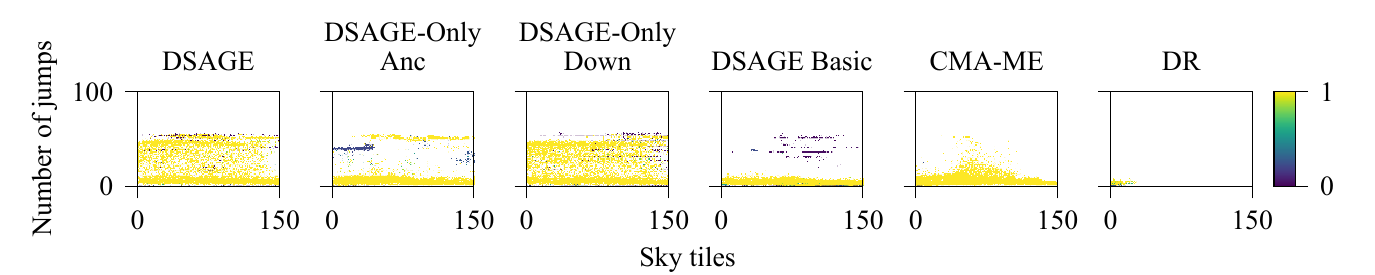}
    \caption{Example archives generated by all algorithms in the Mario domain. Compared to \cmame{}, \dsage{} and its variants are more adept at finding levels with high numbers of jumps. The algorithms then differ in the objective values of the levels that have high numbers of jumps: \dsageb{} finds levels with low objective values, so its QD\=/score is low. \dsageod{} finds many levels with high numbers of jumps, but many of these levels have low objective values (hence the dark region in the top right of its archive), leading to a lower QD\=/score than \dsage{}, which primarily finds levels with high objective values. Note that in our experiments, we never observed a level that caused Mario to jump more than 60 times, so the upper portion of all archives is unoccupied.}
    \label{fig:heatmap_figure_mario}
\end{figure}

\fref{fig:heatmap_figure_maze} and \fref{fig:heatmap_figure_mario} show typical archives output by the algorithms in our experiments in the Maze and Mario domains, respectively.

\newpage
\section{Searching for Additional Agent Behaviors}

Here we present example results from different measures in the Maze and Mario domains. By searching for these measures with \dsage{}, we discover environments that elicit a wide range of agent behaviors not presented in our main paper.

\subsection{Maze}

\begin{figure}
    \centering
    \includegraphics[width=\linewidth]{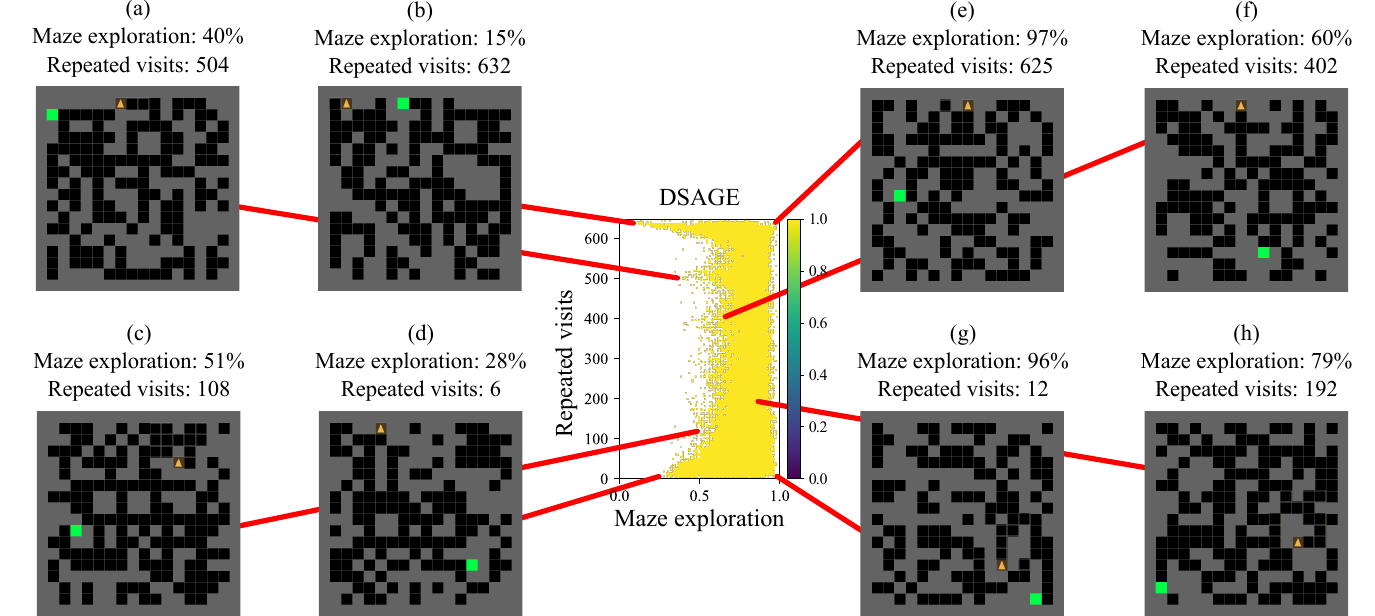}
    \caption{A \dsage{} run in the Maze domain where the measures are (1) the fraction of reachable cells that the agent has visited (termed the ``Maze exploration''; range [0, 1]) and (2) the number of times that the agent visits a cell it has already visited (termed the ``Repeated visits''; range [0, 648]). Note that both of these measures are agent-based.
    In (a) and (b), the agent becomes stuck in a small portion of the maze, leading to many repeated visits but low maze exploration. Notably, the agent observes the goal multiple times in (b), but it never figures out how to go around the large wall which blocks it.
    In (c), the agent gets stuck in several traps (leading to repeated visits) but eventually makes its way to the goal.
    In (d), the agent heads directly to the goal, so it does not explore the maze much, and the only repeated visits it makes come from turning (when the agent turns, it stays in the same cell, which counts as a repeated visit).
    In (e) and (f), the agent visits multiple parts of the maze several times and is unable to reach the goal.
    In (g), the agent explores all of the space without revisiting many cells and eventually finds the goal.
    Finally, in (h), the agent has many repeated visits because it gets stuck at the beginning, but afterwards, it explores the rest of the maze and finds the goal.
    Refer to the supplemental material for videos of these agents.}
    \label{fig:maze_repeated_exploration}
\end{figure}

\begin{figure}
    \centering
    \includegraphics[width=\linewidth]{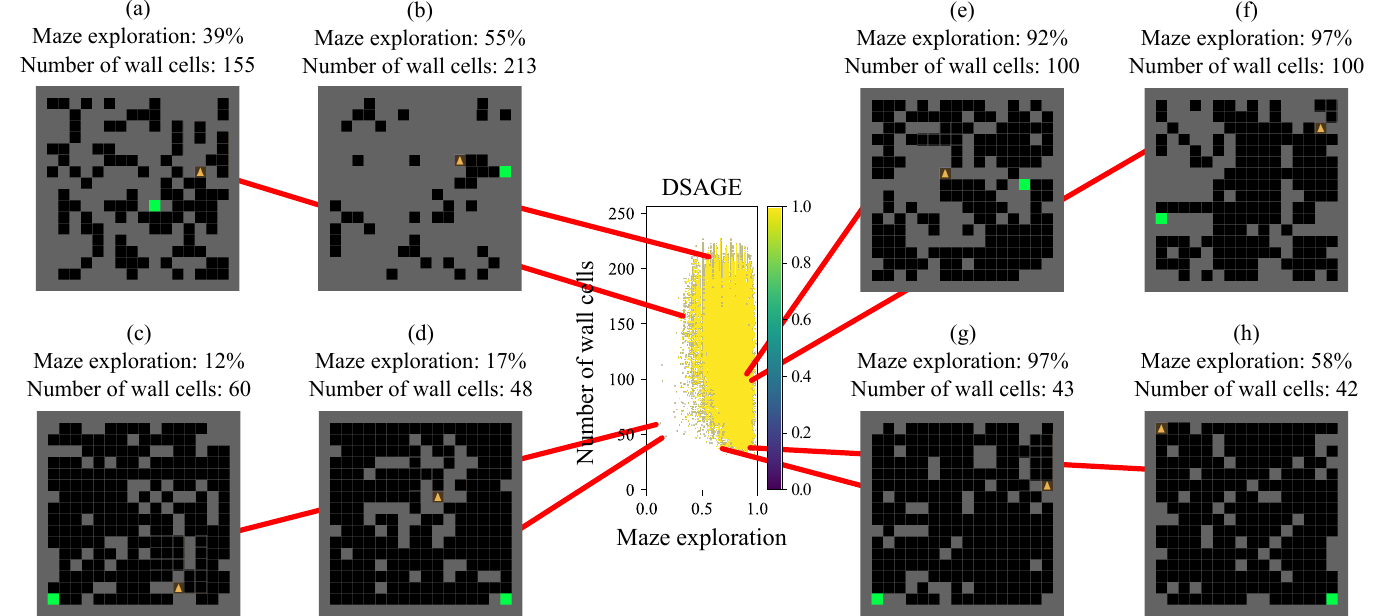}
    \caption{A \dsage{} run in the Maze domain where the measures are (1) the fraction of reachable cells that the agent has visited (termed the ``Maze exploration''; range [0, 1]) and (2) the number of wall cells in the maze (range [0, 256]). (a) and (b) are mazes where the wall cells define a straightforward path for the agent to follow to the goal, resulting in low maze exploration. In (c), the agent goes in circles in the bottom right corner, resulting in low exploration. (d) has a similar number of wall cells to (c), but the agent here is able to quickly find the goal, which also results in low exploration. (e) and (f) are two levels that are similar in terms of both measures yet have very different structures -- in particular, (f) has a much larger reachable space for the agent to explore. In (g), the agent spends all its time exploring even though there are relatively few wall cells blocking its path. Finally, (h) has a similar number of wall cells as (g), but the agent heads almost directly to the goal. Refer to the supplemental material for videos of these agents.}
    \label{fig:maze_walls_exploration}
\end{figure}

\begin{figure}
    \centering
    \includegraphics[width=\linewidth]{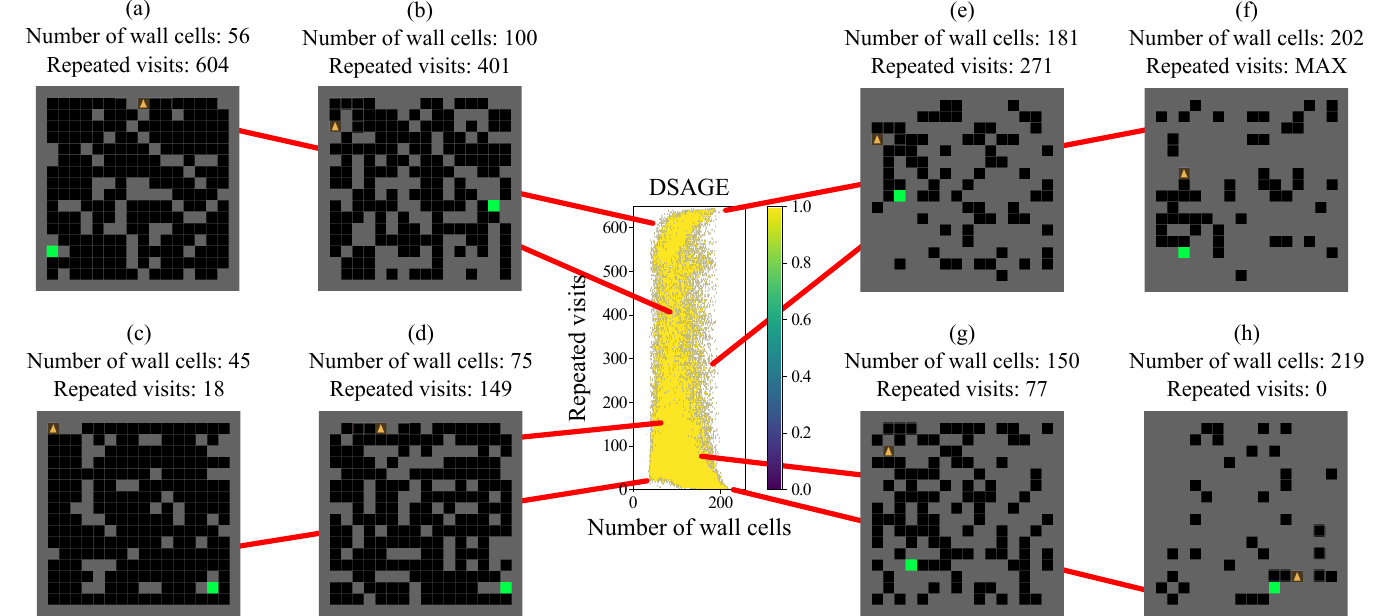}
    \caption{A \dsage{} run in the Maze domain where the measures are (1) the number of wall cells in the maze (range [0, 256]) and (2) the number of times that the agent visits a cell it has already visited (termed the ``Repeated visits''; range [0, 648]). In (a), the agent gets stuck in the top right corner since it is surrounded by walls with only one path out, so it has many repeated visits. In (b), the agent repeatedly goes around the maze and even sees the goal several times, but it usually does not reach the goal. (c) and (d) are relatively easy for the agent --- since it finds the path quickly, it does not repeat many visits. (e) and (f) are cases where the agent gets stuck going in loops even though it is right next to the goal, which leads to many repeated visits. In (g), the agent makes several loops but eventually finds the goal. Finally, in (h), the agent goes directly to the goal, so it never repeats any visits. Refer to the supplemental material for videos of these agents.}
    \label{fig:maze_walls_repeated}
\end{figure}

\fref{fig:maze_repeated_exploration}, \ref{fig:maze_walls_exploration}, and \ref{fig:maze_walls_repeated} show results from \dsage{} runs in the Maze domain with different measures.

\subsection{Mario}

\begin{figure}[h]
    \centering
    \includegraphics[width=\linewidth]{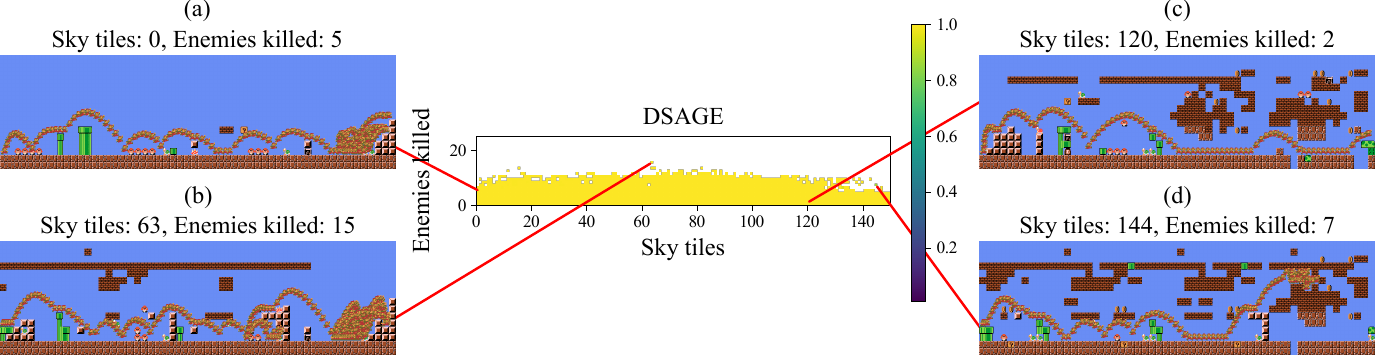}
    \caption{A \dsage{} run in the Mario domain where the measures are (1) the number of sky tiles (exactly as in the main paper) and (2) the number of enemies Mario kills (range [0, 25]). Note that in (b), Mario kills many enemies because Mario repeatedly jumps on the bullets fired by the cannon at the end of the level. In (d), even though Mario kills multiple enemies, Mario cannot complete the level because the sky tiles form an unbreakable barrier. Refer to the supplemental material for videos of these agents.}
    \label{fig:mario_enemies_sky_tiles}
\end{figure}

\begin{figure}[h]
    \centering
    \includegraphics[width=\linewidth]{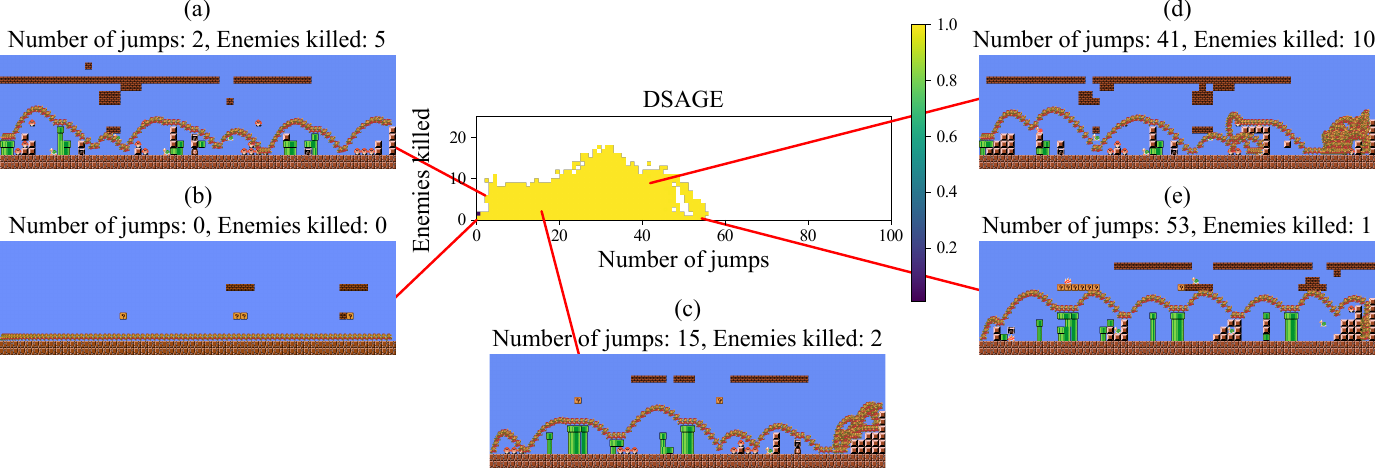}
    \caption{A \dsage{} run in the Mario domain where the measures are (1) the number of times that Mario jumps (range [0, 100]) and (2) the number of enemies that Mario kills (range [0, 25]). Similar to the levels from our earlier experiment (\fref{fig:mario_heatmap}(c)), levels (c), (d), and (e) here have a ``staircase trap'' at the end which causes Mario to perform many jumps, where different trap structures result in different numbers of jumps. Note that in some environments, there appear to be more jumps than indicated in the measures because Mario bounces whenever Mario lands on and kills an enemy, but these bounces do not count as jumps. Refer to the supplemental material for videos of these agents.}
    \label{fig:mario_jumps_enemies}
\end{figure}

\fref{fig:mario_enemies_sky_tiles} and \ref{fig:mario_jumps_enemies} show results from \dsage{} runs in the Mario domain with different measures.

\end{document}